\documentclass[11pt,a4paper]{article}
\usepackage[hyperref]{emnlp2020}
\usepackage{times}
\usepackage{latexsym}
\usepackage{amssymb}

\usepackage{microtype}
\usepackage{graphicx}
\usepackage{subfigure}
\usepackage{booktabs} 
\usepackage{amsmath}
\usepackage{soul}
\usepackage{xspace}
\usepackage{wrapfig,lipsum,booktabs}
\usepackage{algorithm,algpseudocode}
\usepackage{longtable}
\usepackage{wrapfig}

\newcommand{\blue}[1]{{\bf \textcolor{blue}{#1}}}

\usepackage{microtype}

\aclfinalcopy 

\usepackage{array}
\newcolumntype{H}{>{\setbox0=\hbox\bgroup}c<{\egroup}@{}}

\title{\textsc{Pointer}: Constrained Progressive Text Generation via Insertion-based \\ Generative Pre-training}

\author{Yizhe Zhang$^{1}$\Thanks{~These authors contributed equally to this work.}~ \quad\quad Guoyin Wang$^{2}$\footnotemark[1]~~\Thanks{~Work was done while Guoyin was at Microsoft.}~ \quad\quad Chunyuan Li$^{1}$ \\ \textbf{Zhe Gan}$^{1}$\quad\quad \textbf{Chris Brockett}$^{1}$\quad\quad  \textbf{Bill Dolan$^{1}$}\\
  $^{1}$Microsoft Research, Redmond, WA, USA \\
  $^{2}$Amazon Alexa AI, Seattle, WA, USA \\
  {\small \tt \{yizzhang,chunyl,zhe.gan,chrisbkt,billdol\}@microsoft.com, guoyinwang.duke@gmail.com}}

\date{}

\begin{document}
\maketitle
\begin{abstract}
Large-scale pre-trained language models, such as BERT and GPT-2, have achieved excellent performance in language representation learning and free-form text generation. However, these models cannot be directly employed to generate text under specified lexical constraints. To address this challenge, we present \textsc{Pointer}\footnote{\textbf{P}r\textbf{O}gressive \textbf{IN}sertion-based \textbf{T}ransform\textbf{ER}}, a simple yet novel insertion-based approach for hard-constrained text generation. The proposed method operates by progressively inserting new tokens between existing tokens in a parallel manner. This procedure is recursively applied until a sequence is completed. The resulting coarse-to-fine hierarchy makes the generation process intuitive and interpretable. 
We pre-train our model with the proposed progressive insertion-based objective on a 12GB Wikipedia dataset, and fine-tune it on downstream hard-constrained generation tasks. 
Non-autoregressive decoding yields an empirically logarithmic time complexity during inference time. Experimental results on both News and Yelp datasets demonstrate that \textsc{Pointer} achieves state-of-the-art performance on constrained text generation.
We released the pre-trained models and the source code to facilitate future research \footnote{https://github.com/dreasysnail/POINTER}.

\end{abstract}

\section{Introduction}
Real-world editorial assistant applications must often generate text under specified lexical constraints, for example, convert a meeting note with key phrases into a concrete meeting summary, recast a user-input search query as a fluent sentence, generate a conversational response using grounding facts~\cite{mou2016sequence}, or create a story using a pre-specified set of keywords~\cite{Fan2018HierarchicalNS, yao2019plan, donahue2020enabling}. 

Generating text under specific lexical constraints is challenging. 
\emph{Constrained} text generation broadly falls into two categories, depending on whether inclusion of specified keywords in the output is mandatory. 
In \textit{soft-constrained} 
generation \cite{Qin2019CMR, tang2019target}, keyword-text pairs are typically first constructed (sometimes along with other conditioning information), and a conditional text generation model is trained to capture their co-occurrence, so that the model learns to incorporate the constrained keywords into the generated text. 
While \textit{soft-constrained} models are easy to design, even remedied by soft enforcing algorithms such as attention and copy mechanisms  ~\cite{bahdanau2014neural,gu2016incorporating, chen2019improving}, keywords are still apt to be lost during generation,  
 especially when multiple weakly correlated keywords must be included.



\begin{table*}[t!]\centering
	\footnotesize
    \begin{tabular}{|c|l|}
        \hline
         Stage & Generated text sequence  \\
         \hline
         0 ($X^0$) &  {\small  sources sees structure perfectly }\\
         1 ($X^1$)&  {\small  sources \blue{company} sees \blue{change} structure perfectly \blue{legal} } \\
         2 ($X^2$)&  {\small  sources \blue{suggested} company sees \blue{reason}  change \blue{tax} structure \blue{which} perfectly legal \blue{.}} \\
         3 ($X^3$)&  {\small \blue{my} sources \blue{have} suggested \blue{the} company sees \blue{no} reason \blue{to} change \blue{its} tax structure \blue{,} which \blue{are} perfectly legal .}\\
         4 ($X^4$)&  {\small my sources have suggested the company sees no reason to change its tax  structure , which are perfectly legal .}\\
        \hline
    \end{tabular}
    \caption{Example of the progressive generation process with multiple stages from the \textsc{Pointer} model. Words in \blue{blue} indicate newly generated words at the current stage. $X^i$ denotes the generated partial sentence at Stage $i$. $X^4$ and $X^3$ are the same indicates the end of the generation process. Interestingly, our method allows informative words (\emph{e.g.}, \emph{company, change}) generated before the non-informative words (\emph{e.g.}, \emph{the, to}) generated at the end. }
    \label{tab:stage_seq}
    \vspace{-3mm}
\end{table*}

\textit{Hard-constrained} generation~\cite{hokamp2017lcd,post2018fast,hu2019improved,miao2019cgmh,welleck2019non}, on the other hand, requires that all the lexical constraints be present in the output sentence. This approach typically involves sophisticated design of network architectures.  \citet{hokamp2017lcd}
construct a lexical-constrained grid beam search decoding algorithm to incorporate constraints. 
However, \citet{hu2019improved} observe that a naive implementation of this algorithm has a high running time complexity. 
\citet{miao2019cgmh} introduces a sampling-based conditional generation method, where the constraints are first placed in a template, then words in a random position are either inserted, deleted or updated under a Metropolis-Hastings-like scheme.  
However, 
individually sampling each token results in slow convergence, as the joint distribution of all the tokens in a sentence is highly correlated.
\citet{welleck2019non} propose a tree-based text generation scheme, where a token is first generated in an arbitrary position, and then the model recursively generates words to its left and right,  
yielding a binary tree. However, the constructed tree may not reflect the progressive hierarchy/granularity from high-level concepts to low-level details. Further, the time complexity of generating a sentence is $\mathcal{O}(n)$, like standard auto-regressive methods.


Motivated by the above, we propose a novel non-autoregressive model for hard-constrained text generation, called \textsc{Pointer} (\textbf{P}r\textbf{O}gressive \textbf{IN}sertion-based \textbf{T}ransform\textbf{ER}). As illustrated in Table~\ref{tab:stage_seq},
generation of words in \textsc{Pointer} is \textit{progressive}, and \textit{iterative}. Given lexical constraints, \textsc{Pointer} first generates high-level words (\emph{e.g.}, nouns, verbs and adjectives) that bridge the keyword constraints, then these words are used as pivoting points at which to insert details of finer granularity. This process iterates until a sentence is finally completed by adding the least informative words (typically pronouns and prepositions). 

Due to the resemblance to the masked language modeling (MLM) objective, BERT\cite{devlin2019bert} can be naturally utilized for initialization. 
 Further, 
we perform large-scale pre-training on a large Wikipedia corpus to obtain a pre-trained \textsc{Pointer} model that which can be readily fine-tuned on specific downstream tasks. 


The main contributions of this paper are summarized as follows. ($i$) We present \textsc{Pointer}, a novel insertion-based Transformer model for hard-constrained text generation. Compared with previous work, \textsc{Pointer} allows long-term control over generation due to the top-down progressive structure, and enjoys a significant reduction over emperical time complexity from $\mathcal{O}(n)$ to $\mathcal{O}(\log n)$ at best.
 ($ii$) Large-scale pre-training and novel beam search algorithms are proposed to further boost performance. 
 ($iii$) We develop a novel beam search algorithm customized to our approach, further improving the generation quality.
 ($iv$) Experiments on several datasets across different domains (including News and Yelp) demonstrates the superiority of \textsc{Pointer} over strong baselines. Our approach is simple to understand and implement, yet powerful, and can be leveraged as a building block for future research. 

%


\section{Related Work}
\paragraph{Language Model Pre-training}
Large-scale pre-trained language models, such as BERT~\cite{devlin2019bert}, RoBERTa~\cite{liu2019roberta}, XLNet~\cite{yang2019xlnet}, Text-to-text Transformer~\cite{2019t5} and ELECTRA~\cite{clark2020electra}, have achieved great success on natural language understanding benchmarks.
GPT-2~\cite{gpt2} first demonstrates great potential for leveraging Transformer models in generating realistic text. MASS~\cite{song2019mass} and BART~\cite{lewis2019bart} propose methods for sequence-to-sequence pre-training. 
UniLM \cite{dong2019unified} unifies the generation and understanding tasks within a single pre-training scheme. DialoGPT~\cite{zhang2019dialogpt} and MEENA~\cite{adiwardana2020towards} focus on open-domain conversations.
CTRL~\cite{keskarCTRL2019} and Grover~~\cite{zellers2019defending} guide text generation with pre-defined control codes.
To the best of our knowledge, ours is the first large-scale pre-training work for hard-constrained text generation.



\vspace{5pt}
\noindent \textbf{Non-autoregressive Generation}\,
Many attempts have been made to use non-autoregressive models for text generation tasks.
 For neural machine translation, the promise of such methods mostly lies in their decoding efficiency.  
For example, \citet{gu2017non} employs a non-autoregressive decoder that generates all the tokens simultaneously. 
Generation can be further refined with a post-processing step to remedy the conditional independence of the parallel decoding process~\cite{lee2018deterministic,ghazvininejad2019mask,ma2019flowseq, sun2019fast, kasai2020parallel}. 
Deconvolutional decoders \cite{zhang2017deconvolutional, wu2019pay} have also been studied for title generation and machine translation. 
The Insertion Transformer \cite{stern2019insertion,gu2019insertion, chan2019kermit} is a partially autoregressive model that predicts both insertion positions and tokens, and is trained to maximize the entropy over all valid insertions, providing fast inference while maintaining good performance. 
Our \textsc{Pointer} model hybridizes the BERT and Insertion Transformer models, inheriting the advantages of both, and generates text in a progressive coarse-to-fine manner.

\vspace{-1mm}
\section{Method}
\vspace{-1mm}


\subsection{Model Overview} \label{sec:model_overview}
Let $X=\{x_0, x_1,\cdots,x_T\}$ denote a sequence of discrete tokens,
where each token $x_t\in V$, and $V$ is a finite vocabulary set. 
For the hard-constrained text generation task, the goal is to generate a complete text sequence $X$, given a set of key words $\hat{X}$ as constraints, where the key words have to be exactly included in the final generated sequence with the same order. 

Let us denote the lexical constraints as $X^0=\hat{X}$.  The generation procedure of our method can be formulated as a (progressive) sequence of $K$ stages: $S=\{X^0, X^1, \cdots, X^{K-1}, X^K\}$, such that for each $k\in \{1,\ldots, K\}$, $X^{k-1}$ is a sub-sequence of $X^{k}$. The following stage can be perceived as a finer-resolution text sequence compared to the preceding stage.  $X^K$ is the final generation, under the condition that the iterative procedure is converged (\emph{i.e.}, $X^{K-1}=X^K$). 

Table \ref{tab:stage_seq} shows an example of our progressive text generation process. 
Starting from the lexical constraints ($X_0$), at each stage, the algorithm inserts tokens progressively to formulate the target sequence. At each step, at most one new token can be generated between two existing tokens. 
Formally, we propose to factorize the distribution according to the \textit{importance} (defined later) of each token:
\begin{align}
    p(X) &=p(X^0) \prod_{k=1}^K p(X^k|X^{k-1}) \label{eq:progressrive} 
\end{align}
where $p(X^k|X^{k-1})=\prod_{x \in X^k-X^{k-1}} p(x|X^{k-1})$. The more important tokens that form the skeleton of the sentence, such as nouns and verbs, appear in earlier stages, and the auxiliary tokens, such as articles and prepositions, are generated at the later stages. In contrast,
 the autoregressive model factorizes the joint distribution of $X$ in a standard left-to-right manner, \emph{i.e.}, $ p(X) = p(x_0)\prod_{t=1}^T p(x_t|x_{<t})$, ignoring the word importance. 
Though the Insertion Transformer~\cite{stern2019insertion} attempts to implement the progressive generation agenda in \eqref{eq:progressrive}, it does not directly address how to train the model to generate important tokens first. 




\subsection{Data Preparation}
\label{sec:datap}

Designing a loss function so that $(i)$ generating an important token first and $(ii)$ generating more tokens at each stage that would yield a lower loss would be complicated. Instead, we prepare data in a form that eases model training.

The construction of data-instance pairs reverses the generation process.
We construct pairs of text sequences at adjacent stages, \emph{i.e.}, $(X^{k-1},X^k)$, as the model input. Therefore, each training instance $X$ is broken into a consecutive series of pairs: $(X^0,X^1),\cdots,(X^{K-1},X^K)$, where $K$ is the number of such pairs. At each iteration, the algorithm masks out a proportion of existing tokens $X^k$ to yield a sub-sequence $X^{k-1}$, creating a training instance pair $( X^{k-1}, X^k)$. This procedure is iterated until only less than $c$ ($c$ is small) tokens are left. 

Two properties are desired when constructing data instances: ($i$) important tokens should appear in an earlier stage 
, so that the generation follows a progressive manner; ($ii$) the number of stages $K$ is small, thus the generation is fast at inference time.

\vspace{5pt}
\noindent\textbf{Token Importance Scoring}\,
We consider three different schemes to assess the importance score of a token:
term frequency-inverse document frequency (TF-IDF), part-of-speech (POS) tagging, and Yet-Another-Keyword-Extractor (YAKE)~\cite{campos2018yake,campos2020yake}. 
The TF-IDF score provides the uniqueness and local enrichment evaluation of a token at a corpus level. 
POS tagging 
indicates the role of a token at a sequence level.  We explicitly assign noun or verb tokens a higher POS tagging score than tokens from other categories. 
YAKE is a commonly used unsupervised automatic keyword extraction method that relies on statistical features extracted from single documents to select the most important keywords~\cite{campos2020yake}. YAKE is good at extracting common key words, but relatively weak at extracting special nouns (\emph{e.g.}, names), and does not provide any importance level for non-keyword tokens. Therefore, we combine the above three metrics for token importance scoring. Specifically, the overall score $\alpha_t$ of a token $x_t$ is defined as $\alpha_t = \alpha^{\text{TF-IDF}}_t + \alpha^{\text{POS}}_t + \alpha^{\text{YAKE}}_t$, 
where $\alpha^{\text{TF-IDF}}_t$, $\alpha^{\text{POS}}_t$ and $\alpha^{\text{YAKE}}_t$ represent the TF-IDF, POS tagging and YAKE scores (each is rescaled to $[0,1]$), respectively. 

Additionally, stop words are manually assigned a low importance score. If a token appears several times in a sequence, the latter occurrences are assigned a decayed importance score to prevent the model from generating the same token multiple times in one step at inference time. We note that our choice of components of the importance score is heuristic. It would be better to obtain an unbiased/oracle assessment of importance, which we leave for future work.

\vspace{5pt}
\noindent\textbf{DP-based Data Pair Construction}\,
Since we leverage the Insertion-based Transformer, which allows at most one new token to be generated between each two existing tokens, sentence length at most doubles at each iteration. Consequently, the optimal number of iterations $K$ is $\log(T)$, where $T$ is the length of the sequence. Therefore, generation efficiency can be optimized by encouraging more tokens to be discarded during each masking step when preparing the data. However, masking positional interleaving tokens ignores token importance, and thus loses the property of progressive planning from high-level concepts to low-level details at inference time. In practice, sequences generated by such an approach can be less semantically consistent as less important tokens occasionally steer generation towards random content. 

We design an approach to mask the sequence by considering both token importance and efficiency using dynamic programming (DP). To accommodate the nature of insertion-based generation, the masking procedure is under the constraint that no consecutive tokens can be masked at the same stage.
Under such a condition, we score each token and select a subset of tokens that add up to the highest score (all scores are positive). This allows the algorithm to adaptively choose as many high scored tokens as possible to mask. 

Formally, as an integer linear programming problem \cite{richards2002aircraft}, the objective is to find an optimal masking pattern $\Phi = \{\phi_1,\cdots, \phi_T\}$, where $\phi_t \in \{0,1\}$, and $\phi_t=1$ represents discarding the corresponding token $x_t$, and $\phi_t=0$ indicates $x_t$ remains. For a sequence $X^\prime$, the objective can be formulated as: 
\begin{align} \label{eq:seq-gen} 
	\vspace{-3mm}
    \max \sum_{t=1}^{T} \phi_t (\alpha_{\max} - \alpha_t) , \nonumber  \\  
    \text{ s.t.} \quad \phi_t \phi_{t+1} \neq 1\,, \forall t
    \vspace{-3mm}
\end{align}
%


%


where $\alpha_{\max} = \max_t \{\alpha_t\} $. Though solving Eq.~\eqref{eq:seq-gen} is computationally expensive, one can resort to an analogous problem for a solution, the so-called \textit{House Robbery Problem}, a variant of \textit{Maximum Subarray Problem} \cite{bentley1984programming}, where a professional burglar plans to rob houses along a street and tries to maximize the outcome, but cannot break into two adjacent houses without triggering an alarm. This can be solved using dynamic programming \cite{bellman1954theory} (also known as \textit{Kadane's algorithm} \cite{gries1982note}) as shown in Algorithm \ref{alg:dp}. 


\begin{algorithm}[t]
\centering
\footnotesize
\caption{DP-based Data Pair Construction.}
\label{alg:dp}
\begin{algorithmic}[1]
\State{\bfseries Input:}{ A sequence of discrete tokens $X=\{x_1\cdots,x_T\}$ and its corresponding score list $\{\alpha_{\max}-\alpha_1,\cdots, \alpha_{\max}-\alpha_T\}$}
\State{\bfseries Output:}{ Masking pattern $\Phi = \{\phi_1,\cdots, \phi_T\}$}
\State{\bfseries Initialization:} Accumulating scores $s_1 \leftarrow \alpha_{\max}-\alpha_1$ and $s_2 \leftarrow \max(\alpha_{\max}-\alpha_1,\alpha_{\max}-\alpha_2)$; position tracker $p_1 \leftarrow -\inf$ and $p_2 \leftarrow -\inf$; $\Phi$ = 0
\While {$t \leq T$}
    \State{$s_t \leftarrow \max(s_{t-2} + \alpha_{\max}-\alpha_t, s_{t-1})$}
    \If{$s_t=s_{t-1}$} $p_t\leftarrow t-1$
    \Else  \, $p_t\leftarrow t-2$
    
    \EndIf
    \State{$t \leftarrow t+1$}

\EndWhile

\If{$s_T=s_{T-1}$} $t \leftarrow T-1$
\Else \, $t \leftarrow T-2$, $\phi_{T} \leftarrow 1$
\EndIf

\While{$t \geq 1$}
    \State{$\phi_{t} \leftarrow 1$,$t \leftarrow p_t$}
\EndWhile
\end{algorithmic}
\end{algorithm}

\vspace{-2mm}
\subsection{Model Training} \label{sec:model_training}

\paragraph{Stage-wise Insertion Prediction} With all the data-instance pairs $( X^{k-1},X^k)$ created as described above as the model input, 
we optimize the following objective: 
\begin{align}
    & \mathcal{L} =- \log p(X^{k}|X^{k-1}) \label{eq:stage_gen} \\ 
    & = -\sum_{x \in X^{+}} \log p(x|\Phi^{k-1},X^{k-1}) p(\Phi^{k-1}|X^{k-1})\,, \nonumber
\end{align}
where $X^{+} \triangleq  X^k - X^{k-1}$, and $\Phi^{k-1}$ denotes an indicator vector in the $k$-th stage, representing whether an insertion operation is applied in a slot.

%

%

\begin{figure}
	\includegraphics[width=0.99\linewidth]{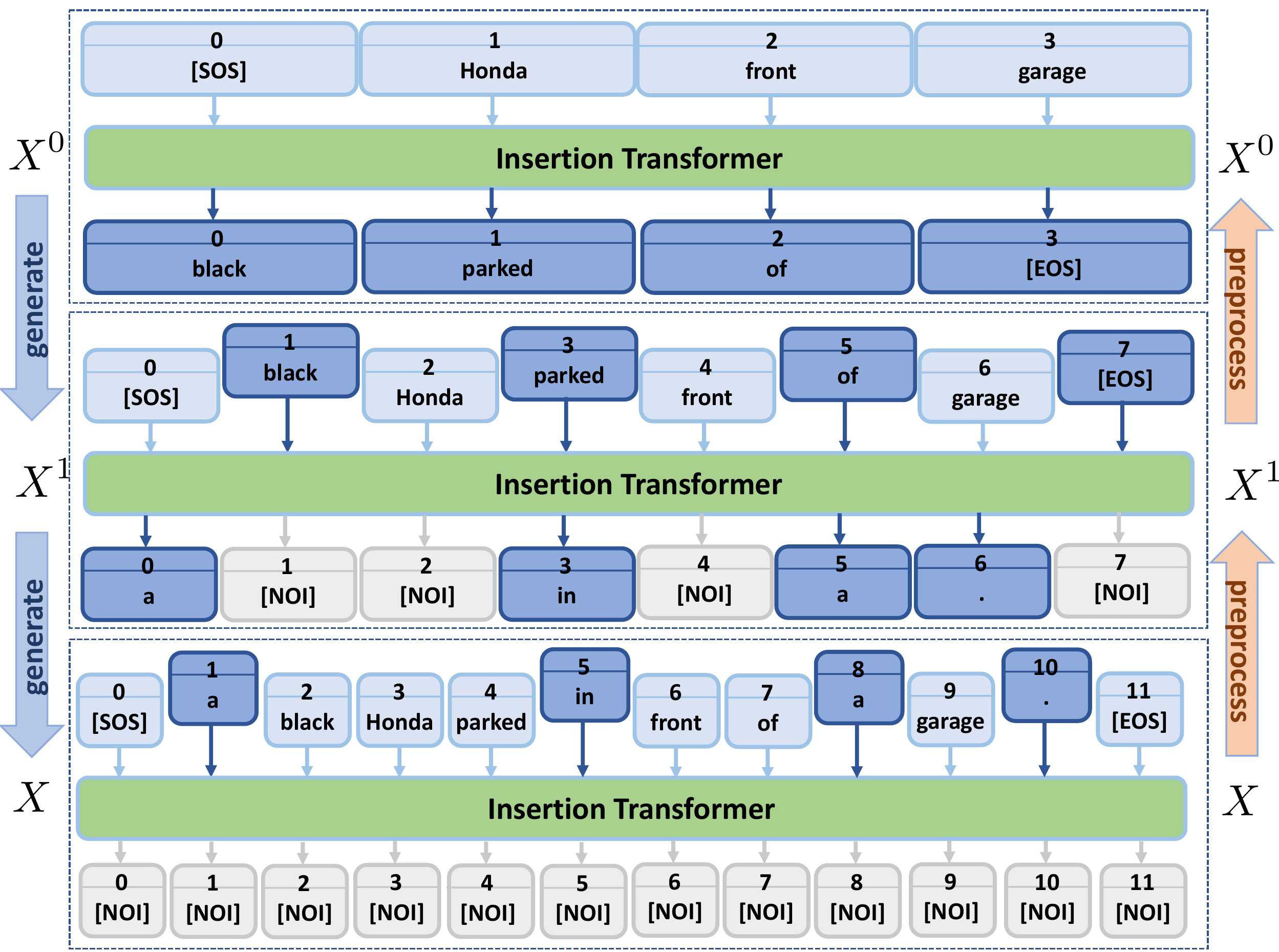}
	\caption{Illustration of the generation process ($X^0\to X$) of the proposed \textsc{Pointer} model. At each stage, the \colorbox{green!20}{Insertion Transformer} module generates either a \colorbox{blue!40}{regular token} or a special \colorbox{gray!10}{$\mathtt{[NOI]}$} token for each gap between two \colorbox{blue!8}{existing tokens}. The generation stops when all the gaps predict $\mathtt{[NOI]}$. The data preparation process reverses the above generative process.} 
	\label{fig:Pointer}
	\vspace{-5mm}
\end{figure}

As illustrated in Figure~\ref{fig:Pointer}, while the MLM objective in BERT only predicts the token of a masked placeholder, our objective comprises both ($i$) likelihood of an insertion indicator for each slot (between two existing tokens), and ($ii$) the likelihood of each new token conditioning on the activated slot. 
To handle this case, we expand the vocabulary with a special no-insertion token $\mathtt{[NOI]}$. During inference time, the model can predict either a token from the vocabulary to insert, or an $\mathtt{[NOI]}$ token indicating no new token will be inserted at a certain slot at the current stage. 
By utilizing this special token, the two objectives are merged. 
Note that the same insertion transformer module is re-used at different stages. We empirically observed that the model can learn to insert different words at different stages; it presumably learns from the “completion level” (how discontinuous the context is) of the current context sequence to roughly estimate the progress up to that point.

During inference time, once in a stage ($X^{k}$), all the slots predict $\mathtt{[NOI]}$ for the next stage, the generation procedure is converged and $X^k$ is the final output sequence.
Note that to account for this final stage $X^k$, during data preparation we incorporate an $(X, N)$ pair for each sentence in the training data, where $N$ denotes a sequence of $\mathtt{[NOI]}$ with the same length of $X$. 
To enable the model to insert at the beginning and end of the sequence, an $\mathtt{[SOS]}$ token and an $\mathtt{[EOS]}$ token are added in the beginning and at the end of each sentence, respectively.



In light of the similarity with the MLM objective, we use BERT model to initialize the Insertion Transformer module. 

\vspace{5pt}
\noindent\textbf{Large-scale Pre-training}\,
In order to provide a general large-scale pretrained model that can benefit various downstream tasks with fine-tuning, we train a model on the massive publicly available English Wiki dataset, which covers a wide range of topics. The Wiki dataset is first preprocessed according to Sec.~\ref{sec:datap}. We then initialize the model with BERT, and perform model training on the processed data using our training objective~\eqref{eq:stage_gen}. After pre-training, the model can be used to generate an appropriate sentence with open-domain keyword constraints, in a tone that represents the Wiki style. In order to adapt the pre-trained model to a new domain (\emph{e.g.}, News and Yelp reviews), the pre-trained model is further fine-tuned on new datasets, which empirically demonstrates better performance than training the model on the target domain alone. 

\subsection{Inference} \label{sec:inference}
During inference time, starting from the given lexical constraint $X^0$, the proposed model generates text stage-by-stage using greedy search or top-K sampling \cite{Fan2018HierarchicalNS}, by applying the Insertion Transformer module repeatedly until no additional token is generated.
If a $\mathtt{[NOI]}$ token is generated, it is deleted at the next round.



\vspace{5pt}
\noindent\textbf{Inner-Layer Beam Search}\,
According to \eqref{eq:stage_gen}, all new tokens are simultaneously generated based on the existing tokens at the previous stage. Despite of being fully parallel, like BERT \cite{yang2019xlnet} and NAT \cite{ghazvininejad2019mask, kasai2020parallel}
this approach suffers from a conditional independence problem in which the predicted tokens are conditional-independently generated and are agnostic of each other. This can result in generating repeating or inconsistent new tokens at each generation round.\footnote{For example, from an existing token ``\emph{and}'', the model generates ``\emph{clean and clean}". } 

To address this weak-dependency issue, we perform a modified beam search algorithm for decoding. 
Specifically, at stage $k$, suppose the existing tokens from last stage are $X^{k-1}=\{x^{k-1}_1,\cdots,x^{k-1}_{T_{k-1}}\}$
, where  $T_{k-1}$ is the length of $X^{k-1}$.
For predicting next stage $X^{k}$, there will be $T_{k-1}$ available slots.
A naive approach to perform beam search would be to maintain a priority queue of top $B$ candidate token series predictions when moving from the leftmost slot to the rightmost slot.
At the $t$-th move, the priority queue contains top $B$ sequences for existing predicted tokens: $(s_1^{(b)},\cdots,s_{t-1}^{(b)})$, where $s_i^{(b)}$ denotes the predicted token for the $i$-th slot in the $b$-th ($b\in\{1,\cdots,B\}$) sequence. The model then evaluates the likelihood of each item (including $\mathtt{[NOI]}$) in the vocabulary for the slot $s_t$, by computing the likelihood of 
$(s_1^{(b)},x^{k-1}_1,\cdots,s_{t-1}^{(b)},x^{k-1}_{t-1},s_t,x^{k-1}_{t},\mathtt{[NOI]},\cdots$, $\mathtt{[NOI]}$,$x^{k-1}_{T_{k-1}})$. This is followed by a ranking step to select the top $B$ most likely series among the $VB$ series to grow. However, such a naive approach is expensive, as the runtime complexity takes $\mathcal{O}(TBV)$ evaluations.

Instead, we approximate the search by constraining it in a narrow band. We design a customized beam search algorithm for our model, called \emph{inner-layer beam search} (ILBS). This method applies an approximate local beam search at each iteration to find the optimal stage-wise decoding.
At the $t$-th slot, ILBS first generates top $B$ token candidates by applying one evaluation step based on existing generation. Prediction is limited to these top $B$ token candidates, and thus the beam search procedure as described above is applied on the narrow band of $B$ instead of the full vocabulary $V$. This reduces the computation to $\mathcal{O}(TB^2)$.







\section{Experiments}
We evaluate the \textsc{Pointer} model on constrained text generation over News and Yelp datasets. Details of the datasets and
experimental results are provided in the following sub-sections. The pre-trained models and the source code are available at Github \footnote{https://github.com/dreasysnail/POINTER}. 
\subsection{Experimental Setup}
\paragraph{Datasets and Pre-processing} 
We evaluate our model on two datasets. The \textit{EMNLP2017 WMT News dataset}\footnote{\url{http://www.statmt.org/wmt17/}} contains 268,586 sentences, and we randomly pick 10k sentences as the validation set, and 1k sentences as the test set. 
The \textit{Yelp English review dataset} is from \citet{cho2018towards}, which contains 160k training examples, 10k validation examples and 1k test examples. 
These two datasets vary in sentence length and domain, enabling the assessment of our model in different scenarios. 

The English Wikipedia dataset we used for pre-training is first pre-processed into a set of natural sentences, with maximum sequence length of 64 tokens,  which results in 1.99 million sentences for model training in total (12.6 GB raw text). On average, each sentence contains 27.4 tokens. 
    
For inference, we extract the testing lexical constraints for all the compared methods using the 3rd party extracting tool YAKE\footnote{\url{https://github.com/LIAAD/yake}}. The maximum length of the lexical constraints we used for News and Yelp is set to 4 and 7, respectively, to account the average length for News ($27.9 \approx 4\times2^3$) and Yelp ($50.3 \approx 7\times2^3$), as we would hope the generation can be done within 4 stages.

\begin{table*}[t!]
\scriptsize
\centering
\resizebox{2.05\columnwidth}{!}{%
\begin{tabular}{r H  r H  r | H r  H r | r| H H H r  |r  r | c |c}
	\cmidrule[\heavyrulewidth]{1-18}
	\blue{News} dataset & \multicolumn{4}{c|}{NIST} & \multicolumn{4}{c|}{BLEU} & METEOR & \multicolumn{4}{c|}{Entropy} & \multicolumn{2}{c|}{Dist} & \multicolumn{1}{c|} {PPL.} & \multicolumn{1}{c} {Avg. Len.}  \\ 
	Method & N-1 & N-2 & N-3 & N-4 & B-1 & B-2 & B-3 & B-4 &  & E-1 & E-2 & E-3 & E-4 &  D-1 &D-2 & & \\
	\cmidrule[\heavyrulewidth]{1-18} 
	CGMH & 1.51 & 1.60 & 1.61 & 1.61 & 17.39\% & 7.09\% & 3.26\% & 1.61\% & 12.55\% & 6.15 & 8.83 & 9.34 & 9.32 & \textbf{16.60\%} & \textbf{70.55\%} & 189.1 & 14.29\\
	NMSTG & 2.57 & 2.70 & 2.70 & 2.70 & 30.13\% & 10.67\% & 3.96\% & 1.58\% & 13.56\% & 6.22 & 9.35 & 10.05 & 10.10 & 11.09\% & 65.96\% & 171.0 & 27.85\\
	\cmidrule[\heavyrulewidth]{1-18} 
	Greedy (base) & 2.71 & 2.90 & 2.80 & 2.80 & 30.73\% & 12.13\% & 4.20\% & 1.63\% & 15.66\% & 5.66 & 8.70 & 10.01 & \textbf{10.41} & 5.89\% & 39.42\% & 97.1 & 47.40\\
	Greedy (+Wiki,base) & 2.84 & 3.04 & 3.06 & 3.06 & 33.73\% & 13.01\% & 5.56\% & 2.51\% & \textbf{16.38\%} & 6.14 & 9.19 & 10.09 & 10.22 & 11.10\% & 57.78\% & 56.7 & 31.32\\
	ILBS (+Wiki,base) & 2.97 & 3.20 & 3.22 & 3.22 & 32.69\% & 14.00\% & 6.24\% & 2.99\% & 15.71\% & 6.05 & 8.97 & 9.77 & 9.86 & 13.17\% & 61.22\% & 66.4 & 22.59\\
	Greedy (+Wiki, large) & 3.04 & \textbf{3.28} & \textbf{3.29} & \textbf{3.30} & 35.75\% & \textbf{14.04\%} & 6.10\% & \textbf{3.04\%} & 15.90\% & 6.15 & 9.17 & 9.97 & 10.09 & 12.23\% & 60.86\% & \textbf{54.7} & 27.99\\
	\cmidrule[\heavyrulewidth]{1-18} 
	Human oracle & - & - & - & - & -& - & - & - & - & 6.17 & 9.20 & 9.93 & 10.05 & 11.80\% & 62.44\% & 47.4 & 27.85\\
	\cmidrule[\heavyrulewidth]{1-18}
	\end{tabular}
}
\resizebox{2.05\columnwidth}{!}{%
%
\begin{tabular}{r H  r H  r | H r  H r | r| H H H r  |r  r | c |c}
	\cmidrule[\heavyrulewidth]{1-18}
	\blue{Yelp} dataset & \multicolumn{4}{c|}{NIST} & \multicolumn{4}{c|}{BLEU} & METEOR & \multicolumn{4}{c|}{Entropy} & \multicolumn{2}{c|}{Dist} &  \multicolumn{1}{c|} {PPL.} & \multicolumn{1}{c} {Avg. Len.} \\ 
	Method & N-1 & N-2 & N-3 & N-4 & B-1 & B-2 & B-3 & B-4 &  & E-1 & E-2 & E-3 & E-4 &  D-1 &D-2 & & \\
	\cmidrule[\heavyrulewidth]{1-18} 
    CGMH & 0.44 & 0.50 & 0.51 & 0.51 & 8.90\% & 4.53\% & 2.47\% & 1.45\% & 11.87\% & 5.64 & 8.53 & 9.36 & 9.48 & \textbf{12.18\%} & \textbf{57.10\%} & 207.2 & 16.70\\
    NMSTG & 1.03 & 1.11 & 1.12 & 1.12 & 24.30\% & 10.06\% & 4.14\% & 1.92\% & 13.88\% & 5.38 & 8.77 & 9.91 & 10.09 & 8.39\% & 50.80\% & 326.4 & 27.92\\
	\cmidrule[\heavyrulewidth]{1-18} 
	Greedy (base) & 2.07 & 2.15 & 2.15 & 2.15 & 27.13\% & 11.48\% & 5.25\% & 2.16\% & \textbf{17.12\%} & 5.89 & 9.04 & 10.46 & \textbf{11.00} & 4.19\% & 31.42\% & 99.5 & 87.30\\
	Greedy (+Wiki,base) & 3.02 & 3.27 & 3.30 & 3.30 & 38.25\% & 15.63\% & 6.93\% & 3.32\% & 16.14\% & 5.96 & 9.18 & 10.37 & 10.64 & 7.51\% & 46.12\% & 71.9 & 48.22\\
	ILBS (+Wiki,base) & 3.08 & 3.34 & 3.39 & 3.38 & 39.63\% & 16.68\% & 7.62\% & 3.65\% & 15.57\% & 5.96 & 9.06 & 10.08 & 10.44 & 9.43\% & 50.66\% & 61.0 & 35.18\\
	Greedy (+Wiki, large)  & 3.19 & \textbf{3.49} & \textbf{3.52} & \textbf{3.53} & 39.41\% & \textbf{16.78\%} & 7.71\% & \textbf{3.79\%} & 16.69\% & 5.79 & 8.89 & 10.19 & 10.56 & 6.94\% & 41.2\% & \textbf{55.5} & 48.05\\
	\cmidrule[\heavyrulewidth]{1-18}
	Human oracle & - & - & - & - & -& - & - & - & - & 6.20 & 9.45 & 10.49 & 10.70 & 10.67\% & 52.57\% & 55.4 & 50.36\\
	\cmidrule[\heavyrulewidth]{1-18}
	\end{tabular}
}
	\vspace{-2mm}
\caption{Automatic evaluation results on the News (upper) and Yelp (lower) dataset. ILBS denotes beam search. ``+Wiki'' denotes fine-tuning on the Wiki-pretrained model. ``base/large'' represents the greedy generation from a based(110M)/large(340M) model. ``Human'' represents the held-out human reference.}\label{tab:auto}
\vspace{-2mm}
\end{table*}

\vspace{5pt}
\noindent\textbf{Baselines} \,
We compare our model with two state-of-the-art methods for hard-constrained text generation: ($i$) Non-Monotonic Sequential Text Generation (NMSTG) \cite{welleck2019non}, and ($ii$) Constrained Sentence Generation by Metropolis-Hastings Sampling (CGMH) \cite{miao2019cgmh}. We also compared with an autoregressive soft-constraint baseline\cite{gao2020mixingboard}. 
Note that the Insertion Transformer~\cite{stern2019insertion} focuses on machine translation rather than hard-constrained generation task, and therefore is not considered for comparison. Other methods based on grid beam search typically have long inference time, and they only operate on the inference stage; these are also excluded from comparison. For all compared system, we use the default settings suggested by the authors, the models are trained until the evaluation loss does not decrease. More details are provided in the Appendix.
 


\vspace{5pt}
\noindent\textbf{Experiment Setups}\,
We employ the tokenizer and model architecture from BERT-base and BERT-large models for all the tasks. BERT models are used as our model initialization. Each model is trained until the validation loss is no longer decreasing. We use a learning rate of 3e-5 without any warming-up schedule for all the training procedures. The optimization algorithm is Adam \cite{kingma2014adam}. We pre-train our model on the Wiki dataset for 2-4 epochs, and fine-tune on the News and Yelp datasets for around 10 epochs. 
 
\vspace{5pt}
\noindent\textbf{Evaluation Metrics}\,
Following \citet{zhang2019dialogpt}, we perform automatic evaluation using commonly adopted text generation metrics, including BLEU \cite{papineni2002bleu}, METEOR \cite{lavie2007meteor}, and NIST \cite{doddington2002nist}. 
Following \cite{kann2018sentence}, to assess the coherence of generated sentences, we also report the perplexity over the test set using pre-trained GPT-2 medium (large) model\footnote{\url{https://github.com/openai/gpt-2}}. 
We use Entropy \cite{zhang2018generating} and Dist-n \cite{li2015diversity} to evaluate lexical diversity.

\begin{table}[ht!]
\footnotesize
\begin{tabular}{p{0.4in}|p{2.3in} }
\cmidrule[\heavyrulewidth]{1-2}
Keywords & \textbf{ estate} \textbf{pay} \textbf{stay} \textbf{policy} \\
\cmidrule[\heavyrulewidth]{1-2}
CGMH &  an economic \blue{estate} developer that could \blue{pay} for it is that a \blue{stay} \blue{policy} .\\
\cmidrule[\heavyrulewidth]{1-2}
NMSTG &  as \blue{estate} owners , they cannot \blue{pay} for households for hundreds of middle - income property , buyers \blue{stay} in retail \blue{policy} .\\
\cmidrule[\heavyrulewidth]{1-2}
\textsc{Pointer} (Greedy, base) & if you buy new buildings from real \blue{estate} company, you may have to \blue{pay} down a mortgage and \blue{stay} with the \blue{policy} for financial reasons . \\
\cmidrule[\heavyrulewidth]{1-2}
\textsc{Pointer} (ILBS, base) & but no matter what foreign buyers do , real \blue{estate} agents will have to \blue{pay} a small fee to \blue{stay} consistent with the \blue{policy} . \\
\cmidrule[\heavyrulewidth]{1-2}
\textsc{Pointer} (Greedy, Large) & but it would also be required for \blue{estate} agents , who must \blue{pay} a larger amount of cash but \blue{stay} with the same \blue{policy} for all other assets . \\
\cmidrule[\heavyrulewidth]{1-2} 

\end{tabular}
\vspace{-2mm}
\caption{Generated examples from the News dataset. }\label{tab:news_gen}
\vspace{-6mm}
\end{table}

\begin{table}[ht!]
\footnotesize
\begin{tabular}{p{0.4in}|p{2.3in} }
\cmidrule[\heavyrulewidth]{1-2}

Keywords & \textbf{joint} \textbf{great} \textbf{food} \textbf{great} \textbf{drinks} \textbf{greater} \textbf{staff} \\
\cmidrule[\heavyrulewidth]{1-2}
CGMH &  very cool \blue{joint} with \blue{great} \blue{food} , \blue{great} \blue{drinks} and even \blue{greater} \blue{staff} . ! .\\
\cmidrule[\heavyrulewidth]{1-2}
NMSTG &  awesome \blue{joint} . \blue{great} service. \blue{great} \blue{food} great \blue{drinks}. good to \blue{greater} and great \blue{staff}!\\
\cmidrule[\heavyrulewidth]{1-2}
\textsc{Pointer} (Greedy, base) & my favorite local \blue{joint} around old town. \blue{great} atmosphere, amazing \blue{food}, delicious and delicious coffee, \blue{great} wine selection and delicious cold \blue{drinks}, oh and maybe even a \blue{greater} patio space and energetic front desk \blue{staff}. \\
\cmidrule[\heavyrulewidth]{1-2}
\textsc{Pointer} (ILBS, base) & the best breakfast \blue{joint} in charlotte . \blue{great} service and amazing \blue{food} . they have \blue{great} selection of \blue{drinks} that suits the \blue{greater} aesthetic of the \blue{staff} . \\
\cmidrule[\heavyrulewidth]{1-2}
\textsc{Pointer} (Greedy, Large) & this is the new modern  breakfast \blue{joint} to be found around the area . \blue{great} atmosphere , central location and excellent \blue{food} . nice variety of selections . \blue{great} selection of local craft beers , good \blue{drinks} . quite cheap unless you ask for \blue{greater} price . very friendly patio and fun \blue{staff} . love it ! \\
\bottomrule
\end{tabular}
\vspace{-2mm}
\caption{Generated examples from the Yelp dataset.}\label{tab:yelp_gen}
\vspace{-6mm}
\end{table}



\subsection{Experimental Results}
\noindent\textbf{News Generation} \,
We first conduct experiments on the News dataset to generate sentences from 4 lexical constraints. Quantitative results are summarized in Table~\ref{tab:auto} (upper). Some qualitative examples including the progressive generations at each stage are provided in Table~\ref{tab:news_gen} and Appendix~\ref{app:news}.
\textsc{Pointer} is able to take full advantage of BERT initialization and Wiki pre-training to improve relevance scores (NIST, BLEU and METEOR). Leveraging the ILBS or using a larger model further improves most automatic metrics we evaluated \footnote{The ILBS for larger models performs similarly to greedy decoding, and thus is omitted from comparison}. For diversity scores, as CGMH is a sampling-based method in nature, it achieves the highest Dist-n scores (even surpasses human score). We observed that the length of generated sentences, the diversity scores and the GPT-2 perplexity from \textsc{Pointer} are close to human oracle. 

\begin{table*}[ht!]
\footnotesize
\centering

\begin{tabular}{r r | r| r l|r r | r| r l}
\cmidrule[\heavyrulewidth]{1-10}

 \multicolumn{10}{c}{\textbf{Semantics}: \textit{A and B, which is more semantically meaningful and consistent?}}\\
 \cmidrule[\heavyrulewidth]{1-10} 
  \multicolumn{5}{c|}{News dataset} & \multicolumn{5}{c}{Yelp dataset} \\
\cmidrule[\heavyrulewidth]{1-10} 
\multicolumn{2}{c|}{System A} & Neutral & \multicolumn{2}{c|}{System B} & \multicolumn{2}{c|}{System A} & Neutral & \multicolumn{2}{c}{System B} \\ 
\cmidrule[\heavyrulewidth]{1-10}
\cmidrule{1-10}
\textsc{Pointer}(base)& \bf{60.9}\% & 17.4\% & 21.8\% & CGMH &
\textsc{Pointer}(base)& \bf{59.8}\% & 17.3\% & 23.0\% & CGMH
\\
\textsc{Pointer}(base)& \bf{55.2}\% & 21.7\% & 23.1\% & NMSTG &
\textsc{Pointer}(base)& \bf{57.5}\% & 23.0\% & 19.6\% & NMSTG
\\
\cmidrule{1-10} 
\textsc{Pointer}(base)& 21.7\% & 21.4\% & \textbf{56.9}\% & Human &
\textsc{Pointer}(base)& 26.8\% & 25.9\% & \textbf{47.3}\% & Human \\
\cmidrule[\heavyrulewidth]{1-10}
%
%
\cmidrule[\heavyrulewidth]{1-10}

 \multicolumn{10}{c}{\textbf{Fluency}: \textit{A and B, which is more grammatical and fluent?}}\\
 \cmidrule[\heavyrulewidth]{1-10} 
  \multicolumn{5}{c|}{News dataset} & \multicolumn{5}{c}{Yelp dataset} \\
\cmidrule[\heavyrulewidth]{1-10} 
\multicolumn{2}{c|}{System A} & Neutral & \multicolumn{2}{c|}{System B} & \multicolumn{2}{c|}{System A} & Neutral & \multicolumn{2}{c}{System B} \\ 
\cmidrule[\heavyrulewidth]{1-10}
\cmidrule{1-10}
\textsc{Pointer}(base)& \bf{57.7}\% & 19.9\% & 22.4\% & CGMH &
\textsc{Pointer}(base)& \bf{54.2}\% & 20.0\% & 25.8\% & CGMH
\\
\textsc{Pointer}(base)& \bf{52.7}\% & 24.1\% & 23.2\% & NMSTG &
\textsc{Pointer}(base)& \bf{59.0}\% & 22.8\% & 18.2\% & NMSTG
\\
\cmidrule{1-10} 
\textsc{Pointer}(base)& 16.6\% & 20.0\% & \textbf{63.4}\% & Human &
\textsc{Pointer}(base)& 24.0\% & 26.1\% & \textbf{49.9}\% & Human
\\
\cmidrule[\heavyrulewidth]{1-10}
%
%
\cmidrule[\heavyrulewidth]{1-10}

 \multicolumn{10}{c}{\textbf{Informativeness}: \textit{A and B, which is more informative?}}\\
 \cmidrule[\heavyrulewidth]{1-10} 
  \multicolumn{5}{c|}{News dataset} & \multicolumn{5}{c}{Yelp dataset} \\
\cmidrule[\heavyrulewidth]{1-10} 
\multicolumn{2}{c|}{System A} & Neutral & \multicolumn{2}{c|}{System B} & \multicolumn{2}{c|}{System A} & Neutral & \multicolumn{2}{c}{System B} \\ 
\cmidrule[\heavyrulewidth]{1-10}
\cmidrule{1-10}
\textsc{Pointer}(base)& \bf{70.4}\% & 12.8\% & 16.8 \% & CGMH &
\textsc{Pointer}(base)& \bf{69.9}\% & 10.9\% & 19.3 \% & CGMH\\
\textsc{Pointer}(base)& \bf{57.7}\% & 18.7\% & 23.6\% & NMSTG &
\textsc{Pointer}(base)& \bf{65.2}\% & 18.1\% & 16.7\% & NMSTG\\
\cmidrule{1-10} 
\textsc{Pointer}(base)& 31.7\% & 19.0\% & \textbf{49.4}\% & Human &
\textsc{Pointer}(base)& 32.8\% & 19.0\% & \textbf{48.2}\% & Human \\
\cmidrule[\heavyrulewidth]{1-10}
\end{tabular}

	\vspace{-2mm}
\caption{{\bf Human Evaluation} on two datasets for semantic consistency, fluency and informativeness, showing preferences (\%) for our \textsc{Pointer}(base) model vis-\`a-vis baselines and real human responses. Numbers in bold indicate the most preferred systems. Differences in mean preferences are statistically significant at $p \leq 0.00001$. 
\vspace{-3mm}
}\label{tab:human_eval}

\end{table*}


\vspace{5pt}
\noindent\textbf{Yelp Generation} \,
We further evaluate our method on the Yelp dataset, where the goal is to generate a long-form text from more constraints. Generating a longer piece of text with more lexical constraints is generally more challenging, since the model needs to capture the long-term dependency structure from the text, and effectively conjure up with a plan to realize the generation. Results of automatic evaluation are provided in Table~\ref{tab:auto} (lower). Generated examples are shown in Table~\ref{tab:yelp_gen} and Appendix~\ref{app:yelp}. Generally, the generation from our model effectively considers all the lexical constraints, and is semantically more coherent and grammatically more fluent, compared with the baseline methods. The automatic evaluation results is generally consistent with the observations from News dataset, with an exception that Dist-n scores is much lower than the human Dist-n scores. 
Compared with greedy approach, at a cost of efficiency, ILBS is typically more concise and contains less repeated information, a defect the greedy approach occasionally suffers (e.g., Table~\ref{tab:yelp_gen}, ``delicious and delicious'').

For both datasets, most of the generations converges with in 4 stages. 
We perform additional experiments on zero-shot generation from the pre-trained model on both datasets, to test the versatility of pre-training. The generated sentences, albeit Wiki-like, are relatively fluent and coherent (see examples in Appendix~\ref{app:news} and \ref{app:yelp}), and yield relatively high relevance scores (see Appendix~\ref{app:wiki} for details). Interestingly, less informative constraints are able to be expanded to coherent sentences. Given the constraint “is to from”, our model generates ``it is oriented to its east, but from the west''.

The autoregressive soft-constraint baseline\cite{gao2020mixingboard} has no guarantee that it will cover all keywords in the given order, thus we omit it in the Table~\ref{tab:auto}. For this baseline, the percentage of keywords that appear in the outputs are 57\% and 43\% for News and Yelp datasets, respectively. With the similar model size (117M), this baseline’s performance is worse than ours approach in automatic metrics for News dataset (BLEU4: $2.99 \to 1.74$; NIST4: $3.22 \to 1.10$; METEOR: $16\% \to 9\%$; DIST2: $61\% \to 58\%$; PPL: $66 \to 84$). The performance gap in Yelp dataset is even larger due to more lexical constraints.


\vspace{5pt}
\noindent\textbf{Human Evaluation}\,
Using a public crowd-sourcing platform (UHRS), we conducted a human evaluation of 400 randomly sampled outputs (out of 1k test set) of CGMH, NMSTG and our base and large models with greedy decoding. Systems were paired and each pair of system outputs was randomly presented (in random order) to 5 crowd-sourced judges , who ranked the outputs pairwise for coherence, informativeness and fluency using a 5-point Likert-like scale. The human evaluation template is provided in Appendix~\ref{app:human}. 
The overall judge preferences for fluency, informativeness and semantic coherence are presented as percentages of the total "vote" in Table~\ref{tab:human_eval}. P-values are all p<0.00001 (line 721), computed using 10000 bootstrap replications. For inter-annotator agreement, Krippendorff's alpha is 0.23 on the News dataset and 0.18 on the Yelp dataset. Despite the noise, the judgments show a strong across-the-board preference for \textsc{Pointer}(base) over the two baseline systems on all categories. A clear preference for the human ground truth over our method is also observed. The base and large models show comparable human judge preferences on the News dataset, while human judges clearly prefer the large model on Yelp data (see Appendix~\ref{app:human_eval_full} for more details).




\vspace{5pt}
\noindent\textbf{Running-time Comparison}\,
One of the motivations of this work is that at each stage the generation can be parallel, leading to a significant reduction in training and inference. 
We compare the model training time and the inference decoding time of all the methods on the Yelp dataset, and summarize the results in  Table~\ref{tab:speed}. The evaluation is based on a single Nvidia V100 GPU. 
Training time for CGMH and \textsc{Pointer} is relatively fast, while NMSTG processes fewer tokens per second since it needs to generate a tree-like structure for each sentence. With respect to inference time, CGMH is slow, as it typically needs hundreds of sampling iterations to decode one sentence. 

We note there is no theoretical guarantee of $\mathcal{O}(\log N)$ time complexity for our method. However, our approach encourages filling as many slots as possible at each stage, which permits enables the model to achieve an empirical $\mathcal{O}(\log N)$ speed. In our experiment 98\% of generations end within 4 stages. 

Note that our method in Table~\ref{tab:speed} uses greedy decoding.
\begin{table}
\centering
\vspace{-2mm}
\small
\begin{tabular}{l|c|c}
\toprule
Model &Training  & Inference  \\
\cmidrule[\heavyrulewidth]{1-3}
CGMH& 4382 toks/s & 33h \\
NMSTG& 357 toks/s & 487s\\
\textsc{Pointer}& 5096 toks/s & 94s \\
\bottomrule
\end{tabular}
\caption{Speed comparison. ``toks/s'' represents tokens per second. Inference time is computed on 1000 test examples. \textsc{Pointer} uses (greedy, base)}
\label{tab:speed}
\vspace{-2mm}
\end{table}
ILBS is around 20 times slower than greedy. The large model is around 3 times slower than the base model.



\section{Conclusion}
We have presented \textsc{Pointer}, a simple yet powerful approach to generating text from a given set of lexical constraints in a non-autoregressive manner. The proposed method leverages a large-scale pre-trained model (such as BERT initialization and our insertion-based pre-training on Wikipedia) to generate text in a progressive manner using an insertion-based Transformer. Both automatic and human evaluation demonstrate the effectiveness of \textsc{Pointer}. In future work, we hope to leverage 
sentence structure, such as the use of constituency parsing, to further enhance the design of the progressive hierarchy. Our model can be also extended to allow inflected/variant forms and arbitrary ordering of given lexical constraints.

\bibliography{igpt}
\bibliographystyle{acl_natbib}

\clearpage
\appendix
\section*{Appendix}

\section{Baseline and Experimental Details}
For NMSTG, we first convert the lexical constraints into a prefix sub-tree, and then sample a sentence to complete the sub-tree. 
We use the default settings suggested by the authors, and use an LSTM with hidden size of 1024 as the text generator, and select the best performed variants (\emph{annealed}) as our baseline. For CGMH, we use their default setting, which uses an LSTM with hidden size of 300, and set the vocabulary size as 50k. Both models are trained until the evaluation loss does not decrease. During inference, we run CGMH for 500 iterations with default hyperparameters. 

For experiment setup, we employ the tokenizer from BERT, and use WordPiece Embeddings \cite{wu2016google} with a 30k token vocabulary for all the tasks. A special no-insertion token $\mathtt{[NOI]}$ is added to the vocabulary. We utilize the BERT-base and BERT-large models with 12 self-attention layers and 768 hidden dimensions as our model initialization. Each model is trained until there is no progress on the validation loss. We use a learning rate of 3e-5 without any warming-up schedule for all the training procedures. The optimization algorithm is Adam \cite{kingma2014adam}. We pre-train our model on the Wiki dataset for 2 epochs, and fine-tune on the News and Yelp datasets for around 10 epochs.


\section{Additional Generated Examples for News Dataset}
\label{app:news}

 We provide two examples on News dataset for how the model progressively generates the sentences in Table~\ref{tab:add_stage_seq}. All the generations are from the \textsc{Pointer} large model using greedy decoding.
\begin{table}[t!]\centering
	\begin{small}
    \begin{tabular}{c|p{2.4in}}
        \cmidrule[\heavyrulewidth]{1-2}
         Stage & Generated text sequence  \\
        \cmidrule[\heavyrulewidth]{1-2}
         0 ($X^0$) & aware negative immediately sites\\
         \cmidrule[\heavyrulewidth]{1-2}
         1 ($X^1$)& \blue{if} aware \blue{posts} negative \blue{should} immediately \blue{any} sites \blue{posts}\\
         \cmidrule[\heavyrulewidth]{1-2}
         2 ($X^2$)& \blue{would} if \blue{user} aware \blue{that} posts \blue{have} negative \blue{impact} should immediately \blue{related} any \blue{these} sites \blue{remove} posts \\
         \cmidrule[\heavyrulewidth]{1-2}
         3 ($X^3$)& \blue{this} would \blue{prefer} if \blue{the} user \blue{is} aware that \blue{the} posts have \blue{a} negative impact \blue{and} should \blue{be} immediately related \blue{to} any \blue{of} these sites \blue{and} remove \blue{those} posts .\\
        \cmidrule[\heavyrulewidth]{1-2}
    \end{tabular}
    
    \begin{tabular}{c|p{2.4in}}
        \cmidrule[\heavyrulewidth]{1-2}
         Stage & Generated text sequence  \\
        \cmidrule[\heavyrulewidth]{1-2}
         0 ($X^0$) & estate pay stay policy\\
         \cmidrule[\heavyrulewidth]{1-2}
         1 ($X^1$)& \blue{also} estate \blue{agents} pay \blue{amount} stay \blue{same} policy assets\\
         \cmidrule[\heavyrulewidth]{1-2}
         2 ($X^2$)& \blue{it} also \blue{required} estate agents \blue{who} pay \blue{same} amount \blue{cash} stay \blue{with} same  policy \blue{all} assets \\
         \cmidrule[\heavyrulewidth]{1-2}
         3 ($X^3$)& \blue{but} it \blue{would} also \blue{be} required \blue{for} estate agents \blue{,} who \blue{must} pay \blue{the} same amount \blue{of} cash \blue{but} stay with \blue{the} same policy \blue{for} all \blue{other} assets . \\
        \cmidrule[\heavyrulewidth]{1-2}
    \end{tabular}    
    \end{small}
    \caption{Example of the progressive generation process with multiple stages from the \textsc{Pointer} model. New additions at each stage are marked as \blue{blue}. }
    \label{tab:add_stage_seq}
\end{table}

In this section, we also provide some additional examples from the 1k news test data.

\begin{table}[H]
\small
\begin{tabular}{p{0.5in}|p{2.2in} }
\cmidrule[\heavyrulewidth]{1-2}
Keywords &  aware negative immediately sites \\
\cmidrule[\heavyrulewidth]{1-2} 
\textsc{Oracle} & where we become \blue{aware} of any accounts that may be \blue{negative} , we \blue{immediately} contact companies such as Instagram , although we have no control over what they allow on their \blue{sites} . \\
\cmidrule[\heavyrulewidth]{1-2}
CGMH &  not even \blue{aware} of \blue{negative} events including video events \blue{immediately} at stations , Facebook \blue{sites}.\\
\cmidrule[\heavyrulewidth]{1-2}
NMSTG &  health providers in a country for England are \blue{aware} of small health systems - and not non - health care but all \blue{negative} is \blue{immediately} treated by heads of businesses and departments in the \blue{sites} .\\
\cmidrule[\heavyrulewidth]{1-2}
\textsc{Pointer} (Greedy, base) & ` if users are \blue{aware} of the \blue{negative} impact of blocking , how can they so \blue{immediately} ban these \blue{sites} ? ' the researchers wrote . \\
\cmidrule[\heavyrulewidth]{1-2}
\textsc{Pointer} (ILBS, base) & if the users are \blue{aware} of or the \blue{negative} messages , they can \blue{immediately} be transferred to other \blue{sites} . \\
\cmidrule[\heavyrulewidth]{1-2}
\textsc{Pointer} (Greedy, Large) & this would prefer if the user is \blue{aware} that the posts have a \blue{negative} impact and should be \blue{immediately} related to any of these \blue{sites} and remove those posts . \\
\cmidrule[\heavyrulewidth]{1-2} 
Wiki zero-shot& he is not \blue{aware} of the \blue{negative} , and will \blue{immediately} go to the positive \blue{sites} . \\
\cmidrule[\heavyrulewidth]{1-2} 
\end{tabular}

\end{table}

\begin{table}[H]
\small
\begin{tabular}{p{0.5in}|p{2.2in} }
\cmidrule[\heavyrulewidth]{1-2}
Keywords &  children fault left charge \\
\cmidrule[\heavyrulewidth]{1-2} 
\textsc{Oracle} & my relationship with my \blue{children} was seriously affected as they were told time and again that everything was my \blue{fault} , they were even \blue{left} ` in \blue{charge} ' of me if my wife went out of the house . \\
\cmidrule[\heavyrulewidth]{1-2}
CGMH &  his two \blue{children} are the rare \blue{fault} that \blue{left} the police \blue{charge}\\
\cmidrule[\heavyrulewidth]{1-2}
NMSTG &  but despite \blue{children} from hospitals to last one by \blue{fault} backing this month , there have arrived as Mr Hunt has been \blue{left} \blue{charge} .\\
\cmidrule[\heavyrulewidth]{1-2}
\textsc{Pointer} (Greedy, base) & but i found that these \blue{children} were not at school however this was not their \blue{fault} , and if so they were \blue{left} without a parent in \blue{charge} . \\
\cmidrule[\heavyrulewidth]{1-2}
\textsc{Pointer} (ILBS, base) & but my lovely wife and \blue{children} consider that it is not our own \blue{fault} and we should not be \blue{left} alone in \blue{charge} . \\
\cmidrule[\heavyrulewidth]{1-2}
\textsc{Pointer} (Greedy, Large) & i said to my \blue{children} : it ' s not his \blue{fault} the parents \blue{left} him ; the parents should be in \blue{charge} of him . \\
\cmidrule[\heavyrulewidth]{1-2} 
Wiki zero-shot& but for the \blue{children} who are not at a \blue{fault} , they are \blue{left} behind on the \blue{charge} . \\
\cmidrule[\heavyrulewidth]{1-2} 
\end{tabular}
\end{table}

\begin{table}[H]
\small

\begin{tabular}{p{0.5in}|p{2.2in} }
\cmidrule[\heavyrulewidth]{1-2}
Keywords &  estate pay stay policy \\
\cmidrule[\heavyrulewidth]{1-2} 
\textsc{Oracle} & how many people on the \blue{estate} does he think will be affected by the new \blue{pay} - to - \blue{stay} \blue{policy} ? \\
\cmidrule[\heavyrulewidth]{1-2}
CGMH &  an economic \blue{estate} developer that could \blue{pay} for it is that a \blue{stay} \blue{policy}\\
\cmidrule[\heavyrulewidth]{1-2}
NMSTG &  as \blue{estate} owners , they cannot \blue{pay} for households for hundreds of middle - income property , buyers \blue{stay} in retail \blue{policy} .\\
\cmidrule[\heavyrulewidth]{1-2}
\textsc{Pointer} (Greedy, base) & if you buy new buildings from real \blue{estate} company, you may have to \blue{pay} down a mortgage and \blue{stay} with the \blue{policy} for financial reasons . \\
\cmidrule[\heavyrulewidth]{1-2}
\textsc{Pointer} (ILBS, base) & but no matter what foreign buyers do , real \blue{estate} agents will have to \blue{pay} a small fee to \blue{stay} consistent with the \blue{policy} . \\
\cmidrule[\heavyrulewidth]{1-2}
\textsc{Pointer} (Greedy, Large) & but it would also be required for \blue{estate} agents , who must \blue{pay} a larger amount of cash but \blue{stay} with the same \blue{policy} for all other assets . \\
\cmidrule[\heavyrulewidth]{1-2} 
Wiki zero-shot& however , his real \blue{estate} agent agreed to \blue{pay} him for the \blue{stay} under the same \blue{policy} . \\
\cmidrule[\heavyrulewidth]{1-2} 
\end{tabular}

\end{table}

\begin{table}[H]
\small

\begin{tabular}{p{0.5in}|p{2.2in} }
\cmidrule[\heavyrulewidth]{1-2}
Keywords &  managers cut costs million \\
\cmidrule[\heavyrulewidth]{1-2} 
\textsc{Oracle} & he was the third of four \blue{managers} sent in to \blue{cut} \blue{costs} and deal with the city ' s \$ 13 \blue{million} deficit . \\
\cmidrule[\heavyrulewidth]{1-2}
CGMH &  the \blue{managers} , who tried to \blue{cut} off their \blue{costs} , added 20 \blue{million} euros\\
\cmidrule[\heavyrulewidth]{1-2}
NMSTG &  business \blue{managers} \blue{cut} demand for more expensive \blue{costs} in 2017 - by October - is around 5 \blue{million} 8 per cent , and has fallen by 0 . 3 per cent in January and 2017 .\\
\cmidrule[\heavyrulewidth]{1-2}
\textsc{Pointer} (Greedy, base) & under one of its general \blue{managers} , the firm had already \blue{cut} its annual operating \blue{costs} from \$ 13 . 5 \blue{million} to six million euros . \\
\cmidrule[\heavyrulewidth]{1-2}
\textsc{Pointer} (ILBS, base) & and last month , the \blue{managers} announced that it had \blue{cut} its operating \blue{costs} by \$ 30 \blue{million} . \\
\cmidrule[\heavyrulewidth]{1-2}
\textsc{Pointer} (Greedy, Large) & the biggest expense is for the \blue{managers} , where it plans to \blue{cut} their annual management \blue{costs} from \$ 18 . 5 \blue{million} to \$ 12 million . \\
\cmidrule[\heavyrulewidth]{1-2} 
Wiki zero-shot& but then he and all of his \blue{managers} agreed to \blue{cut} off all of the operating \blue{costs} by about 1 \blue{million} . \\
\cmidrule[\heavyrulewidth]{1-2} 
\end{tabular}

\end{table}

\begin{table}[H]
\small
\begin{tabular}{p{0.5in}|p{2.2in} }
\cmidrule[\heavyrulewidth]{1-2}
Keywords &  looked report realized wife \\
\cmidrule[\heavyrulewidth]{1-2} 
\textsc{Oracle} & i \blue{looked} at the \blue{report} and saw her name , and that's when I \blue{realized} it was my ex-\blue{wife} . \\
\cmidrule[\heavyrulewidth]{1-2}
CGMH &  he \blue{looked} at the \blue{report} and said he \blue{realized} that if his \blue{wife} Jane\\
\cmidrule[\heavyrulewidth]{1-2}
NMSTG & i \blue{looked} at my \blue{report} about before I \blue{realized} I return to travel holidays but - it doesn ' t haven ' t made anything like my \blue{wife} .\\
\cmidrule[\heavyrulewidth]{1-2}
\textsc{Pointer} (Greedy, base) & when i turned and \blue{looked} at a file \blue{report} from the airport and \blue{realized} it was not my \blue{wife} and daughter . \\
\cmidrule[\heavyrulewidth]{1-2}
\textsc{Pointer} (ILBS, base) & when i turned around and \blue{looked} down at the pictures from the \blue{report} , i \blue{realized} that it was my \blue{wife} . \\
\cmidrule[\heavyrulewidth]{1-2}
\textsc{Pointer} (Greedy, Large) & however , when they \blue{looked} at the details of the \blue{report} about this murder , they quickly \blue{realized} that the suspect was not with his \blue{wife} or his partner . \\
\cmidrule[\heavyrulewidth]{1-2} 
Wiki zero-shot& but when he \blue{looked} up at the \blue{report} , he \blue{realized} that it was not his \blue{wife} . \\
\cmidrule[\heavyrulewidth]{1-2} 
\end{tabular}
\end{table}

\begin{table}[H]
\small

\begin{tabular}{p{0.5in}|p{2.2in} }
\cmidrule[\heavyrulewidth]{1-2}
Keywords &  time claim tax year \\
\cmidrule[\heavyrulewidth]{1-2} 
\textsc{Oracle} & walker says there is still \blue{time} to \blue{claim} this higher protection if you haven ' t already as the deadline is the end of the 2016 / 2017 \blue{tax} \blue{year} . \\
\cmidrule[\heavyrulewidth]{1-2}
CGMH &  " two states , one - \blue{time} voters can \blue{claim} a federal \blue{tax} \blue{year}\\
\cmidrule[\heavyrulewidth]{1-2}
NMSTG & this \blue{time} they had three to \blue{claim} of an equal \blue{tax} and 34 women at which indicated they should leave that over the \blue{year} of 16 .\\
\cmidrule[\heavyrulewidth]{1-2}
\textsc{Pointer} (Greedy, base) & it is the very first \blue{time} in history that trump will ever \blue{claim} over \$ 400 million in federal income \blue{tax} that he had held last \blue{year} , the same report says . \\
\cmidrule[\heavyrulewidth]{1-2}
\textsc{Pointer} (ILBS, base) & is this the very first \blue{time} someone has to \blue{claim} federal income \blue{tax} twice in a single \blue{year} ? \\
\cmidrule[\heavyrulewidth]{1-2}
\textsc{Pointer} (Greedy, Large) & this is not for the first \blue{time} that the scottish government was able to \blue{claim} \blue{tax} cuts of thousands of pounds a \blue{year} to pay . \\
\cmidrule[\heavyrulewidth]{1-2} 
Wiki zero-shot& but at the \blue{time} , the \blue{claim} was that the same sales \blue{tax} that was from the previous fiscal \blue{year} . \\
\cmidrule[\heavyrulewidth]{1-2} 
\end{tabular}

\end{table}

\begin{table}[H]
\small
\begin{tabular}{p{0.5in}|p{2.2in} }
\cmidrule[\heavyrulewidth]{1-2}
Keywords &  model years big drama \\
\cmidrule[\heavyrulewidth]{1-2} 
\textsc{Oracle} & the former \blue{model} said : `` I haven ' t seen him in so many \blue{years} , I can ' t make a \blue{big} \blue{drama} out of it . '' \\
\cmidrule[\heavyrulewidth]{1-2}
CGMH &  the `` \blue{model} '' continues , like many \blue{years} of sexual and \blue{big} \blue{drama} going\\
\cmidrule[\heavyrulewidth]{1-2}
NMSTG & after \blue{model} two \blue{years} and did it like , could we already get bigger than others in a \blue{big} \blue{drama} ? \\
\cmidrule[\heavyrulewidth]{1-2}
\textsc{Pointer} (Greedy, base) & but i am a good role \blue{model} , who has been around for 10 \blue{years} now , and that is a \blue{big} example of what i can do in \blue{drama} on screen . \\
\cmidrule[\heavyrulewidth]{1-2}
\textsc{Pointer} (ILBS, base) & but the young actress and \blue{model} , for 15 \blue{years} , made a very \blue{big} impact on the \blue{drama} . \\
\cmidrule[\heavyrulewidth]{1-2}
\textsc{Pointer} (Greedy, Large) & i have seen the different \blue{model} she recommends of over \blue{years} , but it ' s no \blue{big} change in the \blue{drama} after all . \\
\cmidrule[\heavyrulewidth]{1-2} 
Wiki zero-shot& she was a \blue{model} actress for many \blue{years} and was a \blue{big} star in the \blue{drama} . \\
\cmidrule[\heavyrulewidth]{1-2} 
\end{tabular}
\end{table}

\begin{table}[H]
\small
\begin{tabular}{p{0.5in}|p{2.2in} }
\cmidrule[\heavyrulewidth]{1-2}
Keywords &  made year resolution managed \\
\cmidrule[\heavyrulewidth]{1-2} 
\textsc{Oracle} & i once \blue{made} this my new \blue{year} ' s \blue{resolution} , and it is the only one that I ' ve actually ever \blue{managed} to keep . \\
\cmidrule[\heavyrulewidth]{1-2}
CGMH &  indeed , as he \blue{made} up the previous \blue{year} , the GOP \blue{resolution} was \blue{managed}\\
\cmidrule[\heavyrulewidth]{1-2}
NMSTG & while additional sanctions had been issued last week \blue{made} a \blue{year} from the latest \blue{resolution} , Russia ' s Russian ministers have but have \blue{managed} . \\
\cmidrule[\heavyrulewidth]{1-2}
\textsc{Pointer} (Greedy, base) & no progress has been \blue{made} in syria since the security council started a \blue{year} ago , when a \blue{resolution} expressed confidence that moscow \blue{managed} to save aleppo . \\
\cmidrule[\heavyrulewidth]{1-2}
\textsc{Pointer} (ILBS, base) & and the enormous progress we have \blue{made} over the last \blue{year} is to bring about a \blue{resolution} that has not been \blue{managed} . \\
\cmidrule[\heavyrulewidth]{1-2}
\textsc{Pointer} (Greedy, Large) & the obama administration , which \blue{made} a similar call earlier this \blue{year} and has also voted against a \blue{resolution} to crack down on the funding , \blue{managed} to recover it .  \\
\cmidrule[\heavyrulewidth]{1-2} 
Wiki zero-shot& but despite all the same changes \blue{made} both in both the previous fiscal \blue{year} , and by the un \blue{resolution} itself , only the federal government \blue{managed} ... \\
\cmidrule[\heavyrulewidth]{1-2} 
\end{tabular}
\end{table}

\begin{table}[H]
\small
\begin{tabular}{p{0.5in}|p{2.2in} }
\cmidrule[\heavyrulewidth]{1-2}
Keywords &  club believed centre window \\
\cmidrule[\heavyrulewidth]{1-2} 
\textsc{Oracle} & the \blue{club} are \blue{believed} to be keen on bringing in cover at \blue{centre} - back during the current transfer \blue{window} , with a loan move most likely . \\
\cmidrule[\heavyrulewidth]{1-2}
CGMH &  the \blue{club} has also been \blue{believed} that more than a new \blue{centre} - up \blue{window}\\
\cmidrule[\heavyrulewidth]{1-2}
NMSTG & one \blue{club} \blue{believed} it was not clear that the \blue{centre} would hold place on the \blue{window} until there were no cases that they had heard or had the decision disappeared . \\
\cmidrule[\heavyrulewidth]{1-2}
\textsc{Pointer} (Greedy, base) & he had been talking to the \blue{club} since he is \blue{believed} to have reached the \blue{centre} spot in the queue before the january transfer \blue{window} was suspended . \\
\cmidrule[\heavyrulewidth]{1-2}
\textsc{Pointer} (ILBS, base) & when he left his old \blue{club} , chelsea , he was \blue{believed} to be at the \blue{centre} of the transfer \blue{window} . \\
\cmidrule[\heavyrulewidth]{1-2}
\textsc{Pointer} (Greedy, Large) & the striker has remained at the \blue{club} at the weekend and is increasingly \blue{believed} to be available as a \blue{centre} of the club during the summer transfer \blue{window} until january 2016 . \\
\cmidrule[\heavyrulewidth]{1-2} 
Wiki zero-shot& during his first \blue{club} as manager he was widely \blue{believed}  to be at the \blue{centre} forward in the january transfer \blue{window} . \\
\cmidrule[\heavyrulewidth]{1-2} 
\end{tabular}
\end{table}

\begin{table}[H]
\small
\begin{tabular}{p{0.5in}|p{2.2in} }
\cmidrule[\heavyrulewidth]{1-2}
Keywords &  great past decade city \\
\cmidrule[\heavyrulewidth]{1-2} 
\textsc{Oracle} & it ' s been a \blue{great} time , the \blue{past} \blue{decade} or so , to be the mayor of a major capital \blue{city} . \\
\cmidrule[\heavyrulewidth]{1-2}
CGMH &  the great past decade is that so much of a new home city\\
\cmidrule[\heavyrulewidth]{1-2}
NMSTG & i like to thank you for me and I ' ve wanted it to grow in every \blue{great} \blue{past} \blue{decade} over the \blue{city} , a very amazing time .\\
\cmidrule[\heavyrulewidth]{1-2}
\textsc{Pointer} (Greedy, base) & this is one of the \blue{great} cities that he have visited in the \blue{past} two \blue{decade} , the kansas \blue{city} , missouri , he says . \\
\cmidrule[\heavyrulewidth]{1-2}
\textsc{Pointer} (ILBS, base) & you don ' t feel as \blue{great} as you ' ve been in the \blue{past} \blue{decade} in a major \blue{city} . \\
\cmidrule[\heavyrulewidth]{1-2}
\textsc{Pointer} (Greedy, Large) & there has been a lot of \blue{great} work here in the \blue{past} few years within more than a \blue{decade} , done for the \blue{city} , he says . \\
\cmidrule[\heavyrulewidth]{1-2} 
Wiki zero-shot& there was a \blue{great} success in the \blue{past} during the last \blue{decade} for the \blue{city} . \\
\cmidrule[\heavyrulewidth]{1-2} 
\end{tabular}
\end{table}

\newpage
\section{Additional Generated Examples for Yelp Dataset}
\label{app:yelp}

We provide two examples on Yelp dataset for how the model progressively generates the sentences in Table~\ref{tab:add_stage_yelp_seq}. All the generations are from the \textsc{Pointer} large model using greedy decoding.
 
We also provide some additional examples from the Yelp test set. The results includes keywords, human oracle, CGMH, NMSTG and our models. For our models, we include \textsc{Pointer} base and large models with greedy decoding and base model with ILBS. The large model with ILBS is time consuming so we omit them from the comparison.
\begin{table}[ht!]\centering
	\begin{small}
	 \begin{tabular}{c|p{2.4in}}
        \cmidrule[\heavyrulewidth]{1-2}
         Stage & Generated text sequence  \\
        \cmidrule[\heavyrulewidth]{1-2}
         0 ($X^0$) & delicious love mole rice back\\
         \cmidrule[\heavyrulewidth]{1-2}
         1 ($X^1$)& \blue{restaurant} delicious \blue{authentic} love \blue{dish} mole \blue{beans} rice \blue{definitely} back \blue{!}\\
         \cmidrule[\heavyrulewidth]{1-2}
         2 ($X^2$)& \blue{new} restaurant \blue{so} delicious \blue{fresh} authentic \blue{.} love \blue{mexican} dish \blue{called} mole \blue{with} beans \blue{and} rice \blue{we} definitely \blue{coming} back \blue{more} !   \\
         \cmidrule[\heavyrulewidth]{1-2}
         3 ($X^3$)& \blue{this} new restaurant \blue{is} so delicious \blue{,} fresh \blue{and} authentic \blue{tasting} . \blue{i} love \blue{the} mexican \blue{style} dish \blue{,} called \blue{the} mole \blue{,} with \blue{black} beans \blue{,} and \blue{white} rice \blue{.} we \blue{will} definitely \blue{be} coming back \blue{for} more ! \\
        \cmidrule[\heavyrulewidth]{1-2}
    \end{tabular}  
    \begin{tabular}{c|p{2.4in}}
        \cmidrule[\heavyrulewidth]{1-2}
         Stage & Generated text sequence  \\
        \cmidrule[\heavyrulewidth]{1-2}
         0 ($X^0$) & joint great food great drinks greater staff\\
         \cmidrule[\heavyrulewidth]{1-2}
         1 ($X^1$)& \blue{new} joint \blue{around} great \blue{location} food \blue{variety} great \blue{craft} drinks \blue{unless} greater \blue{friendly} staff \blue{!}\\
         \cmidrule[\heavyrulewidth]{1-2}
         2 ($X^2$)& \blue{is} new \blue{breakfast} joint \blue{be} around \blue{area} great \blue{,} location \blue{excellent} food \blue{nice} variety \blue{selections} great \blue{of} craft \blue{,} drinks \blue{quite} unless \blue{ask} greater \blue{.} friendly \blue{and} staff \blue{love} ! \\
         \cmidrule[\heavyrulewidth]{1-2}
         3 ($X^3$)& \blue{this} is \blue{the} new \blue{modern} breakfast joint \blue{to} be \blue{found} around \blue{the} area \blue{.} great \blue{atmosphere} , \blue{central} location \blue{and} excellent food \blue{.} nice variety \blue{of} selections \blue{.} great \blue{selection} of \blue{local} craft \blue{beers} , \blue{good} drinks . quite \blue{cheap} unless \blue{you} ask \blue{for} greater \blue{price} . \blue{very} friendly \blue{patio} and \blue{fun} staff \blue{.} love \blue{it} ! \\
        \cmidrule[\heavyrulewidth]{1-2}
    \end{tabular}
   \end{small}
    \caption{Example of the progressive generation process with multiple stages from the \textsc{Pointer} model. New additions at each stage are marked as \blue{blue}. }
    \label{tab:add_stage_yelp_seq}
\end{table}

\begin{table}[H]
\begin{tabular}{p{0.5in}|p{2.2in} }
\cmidrule[\heavyrulewidth]{1-2}
Keywords &  service perfect delicious service awesome good place \\
\cmidrule[\heavyrulewidth]{1-2} 
\textsc{Oracle} & yummy excellent \blue{service} . ordered the carne asada medium rare . it was \blue{perfect} . and \blue{delicious} . their customer \blue{service} was \blue{awesome} . they were so friendly and made sure all was \blue{good} . i definitely recommend this \blue{place} . \\
\cmidrule[\heavyrulewidth]{1-2}
CGMH &  great \blue{service} \blue{perfect} food and \blue{delicious} \blue{service} . \blue{awesome} place and \blue{good} \blue{place} !.\\
\cmidrule[\heavyrulewidth]{1-2}
NMSTG &  \blue{service} was \blue{perfect} , \blue{delicious} and great \blue{service} \blue{awesome} service \blue{good} food . this \blue{place} will go back .\\
\cmidrule[\heavyrulewidth]{1-2}
\textsc{Pointer} (Greedy, base) & excellent food , great \blue{service} , really nice atmosphere , \blue{perfect} amount of spring rolls , \blue{delicious} especially the chicken and eel . the \blue{service} was very friendly and the prices are \blue{awesome} too . for a female who loves \blue{good} japanese restaurant , this is definitely your \blue{place} ! \\
\cmidrule[\heavyrulewidth]{1-2}
\textsc{Pointer} (ILBS, base) & from the food to \blue{service} . the foods are \blue{perfect} , they were \blue{delicious} . and \blue{service} is beyond expectation . christina was \blue{awesome} , so many \blue{good} things about this \blue{place} . \\
\cmidrule[\heavyrulewidth]{1-2}
\textsc{Pointer} (Greedy, Large) & absolutely loved the food and very friendly \blue{service} . i had the chicken , it was cooked \blue{perfect} and the seafood pasta was thick and \blue{delicious} and not too heavy though . our \blue{service} guy at the front bar was so \blue{awesome} , he made sure we had a  \blue{good} time . would definitely recommend to try this \blue{place} to anyone ! \\
\cmidrule[\heavyrulewidth]{1-2} 
Wiki zero-shot& he said the \blue{service} was \blue{perfect} , and \blue{delicious} , and the \blue{service} that is \blue{awesome} , and very \blue{good} in its \blue{place} . \\
\cmidrule[\heavyrulewidth]{1-2} 
\end{tabular}
\end{table}

\begin{table}[H]
\begin{tabular}{p{0.5in}|p{2.2in} }
\cmidrule[\heavyrulewidth]{1-2}
Keywords & good drinks love clients tighter great service \\
\cmidrule[\heavyrulewidth]{1-2} 
\textsc{Oracle} & \blue{great} atmosphere , good food and \blue{drinks} . i \blue{love} coming here in the fall to spring to meet with \blue{clients} . their inside is a little small and makes summer a bit \blue{tighter} , but still a \blue{great} staff with excellent \blue{service} . \\
\cmidrule[\heavyrulewidth]{1-2}
CGMH &  \blue{good} \blue{drinks} . i \blue{love} how out \blue{clients} are \blue{tighter} . \blue{great} customer \blue{service} .\\
\cmidrule[\heavyrulewidth]{1-2}
NMSTG &  such \blue{good} place with i love the mushroom \blue{drinks} . the menu they \blue{love} the \blue{clients} . and \blue{tighter} out the menu are \blue{great} \blue{service} .\\
\cmidrule[\heavyrulewidth]{1-2}
\textsc{Pointer} (Greedy, base) & this place is \blue{good} . they have a wide variety of \blue{drinks} . this really fits your taste . \blue{love} the cozy bar that allows \blue{clients} to be able to fit very tightly and \blue{tighter} , better blending with the crowd . \blue{great} coffee , reasonable prices , and friendly \blue{service} ! \\
\cmidrule[\heavyrulewidth]{1-2}
\textsc{Pointer} (ILBS, base) & nice place , with \blue{good} vibe . nice mix of \blue{drinks} and intimate space . what i really \blue{love} about was there were so more mature \blue{clients} , and they can fit in a \blue{tighter} timeline . overall , \blue{great} atmosphere and excellent \blue{service} . \\
\cmidrule[\heavyrulewidth]{1-2}
\textsc{Pointer} (Greedy, Large) & really like this place . has a \blue{good} dj , good atmosphere and cool \blue{drinks} and quite nice lounge area . i \blue{love} this idea of having fun on your \blue{clients} and rubbing your feet to stand up \blue{tighter} than other ones . \blue{great} variety of drinks and pretty quick \blue{service} at the bar ! \\
\cmidrule[\heavyrulewidth]{1-2} 
Wiki zero-shot& she is a \blue{good} at \blue{drinks} , and in \blue{love} for him and all his \blue{clients} , and he enjoys a \blue{tighter} schedule and has a \blue{great} food and a generous \blue{service} . \\
\cmidrule[\heavyrulewidth]{1-2} 
\end{tabular}
\end{table}

\begin{table}[H]
\begin{tabular}{p{0.5in}|p{2.2in} }
\cmidrule[\heavyrulewidth]{1-2}
Keywords & joint great food great drinks greater staff \\
\cmidrule[\heavyrulewidth]{1-2} 
\textsc{Oracle} & apteka is seriously all around the best vegan \blue{joint} in the burgh . \blue{great} \blue{food} , \blue{great} \blue{drinks} , \blue{greater} \blue{staff} . \\
\cmidrule[\heavyrulewidth]{1-2}
CGMH &  very cool \blue{joint} with \blue{great} \blue{food} , \blue{great} \blue{drinks} and even \blue{greater} \blue{staff} . !\\
\cmidrule[\heavyrulewidth]{1-2}
NMSTG &  awesome \blue{joint} . \blue{great} service . \blue{great} \blue{food} great \blue{drinks} . good to \blue{greater} and great \blue{staff} !\\
\cmidrule[\heavyrulewidth]{1-2}
\textsc{Pointer} (Greedy, base) & my favorite local \blue{joint} around old town . \blue{great} atmosphere , amazing \blue{food} , delicious and delicious coffee , \blue{great} wine selection and delicious cold \blue{drinks} , oh and maybe even a \blue{greater} patio space and energetic front desk \blue{staff} . \\
\cmidrule[\heavyrulewidth]{1-2}
\textsc{Pointer} (ILBS, base) & the best breakfast \blue{joint} in charlotte . \blue{great} service and amazing \blue{food} . they have \blue{great} selection of \blue{drinks} that suits the \blue{greater} aesthetic of the \blue{staff} . \\
\cmidrule[\heavyrulewidth]{1-2}
\textsc{Pointer} (Greedy, Large) & this is the new modern  breakfast \blue{joint} to be found around the area . \blue{great} atmosphere , central location and excellent \blue{food} . nice variety of selections . \blue{great} selection of local craft beers , good \blue{drinks} . quite cheap unless you ask for \blue{greater} price . very friendly patio and fun \blue{staff} . love it ! \\
\cmidrule[\heavyrulewidth]{1-2} 
Wiki zero-shot& it is a joint owner of the \blue{great} society of irish \blue{food} , and the \blue{great} britain and soft \blue{drinks} , and the \blue{greater} britain and its \blue{staff} . \\
\cmidrule[\heavyrulewidth]{1-2} 
\end{tabular}
\end{table}

\begin{table}[H]
\begin{tabular}{p{0.5in}|p{2.2in} }
\cmidrule[\heavyrulewidth]{1-2}
Keywords & service polite professional affordable work safe tree \\
\cmidrule[\heavyrulewidth]{1-2} 
\textsc{Oracle} & aron's tree \blue{service} were very \blue{polite} and \blue{professional} . they are very \blue{affordable} . they arrived a little early and got right to \blue{work} . they were quick and \blue{safe} . they cleaned up and hauled out the \blue{tree} trimmings . i highly recommend them . \\
\cmidrule[\heavyrulewidth]{1-2}
CGMH &  excellent customer \blue{service} , \blue{polite} , \blue{professional} , and \blue{affordable} \blue{work} , \blue{safe} bike \blue{tree} .\\
\cmidrule[\heavyrulewidth]{1-2}
NMSTG &  excellent food and \blue{service} and are amazing service and \blue{polite} and \blue{professional} . \blue{affordable} it \blue{work} out \blue{safe} on sun \blue{tree} !\\
\cmidrule[\heavyrulewidth]{1-2}
\textsc{Pointer} (Greedy, base) & amazing customer \blue{service} . so \blue{polite} , and very \blue{professional} , and very \blue{affordable} . such great \blue{work} done at the \blue{safe} end of a \blue{tree} . \\
\cmidrule[\heavyrulewidth]{1-2}
\textsc{Pointer} (ILBS, base) & excellent customer \blue{service} , very \blue{polite} , and very \blue{professional} . honest and \blue{affordable} pricing . i will definitely get the \blue{work} done here for the \blue{safe} parts of my \blue{tree} . \\
\cmidrule[\heavyrulewidth]{1-2}
\textsc{Pointer} (Greedy, Large) & diane provides  customers with great customer \blue{service} . technician mike was very \blue{polite} and helpful . clean facility , very \blue{professional} , and always responsive . quick and \blue{affordable} as well . i had very nice \blue{work} done . we have now found someone \blue{safe} . thank you big two buck \blue{tree} shrub care ! \\
\cmidrule[\heavyrulewidth]{1-2} 
Wiki zero-shot& customer \blue{service} should be more \blue{polite} , and more \blue{professional} , and more \blue{affordable} , and will \blue{work} in a \blue{safe} place under the family \blue{tree} . \\
\cmidrule[\heavyrulewidth]{1-2} 
\end{tabular}
\end{table}

\begin{table}[H]
\begin{tabular}{p{0.5in}|p{2.2in} }
\cmidrule[\heavyrulewidth]{1-2}
Keywords & hesitate give customers chicken rice decent list \\
\cmidrule[\heavyrulewidth]{1-2} 
\textsc{Oracle} & i \blue{hesitate} to \blue{give} them the five stars they deserve because they have a really small dining area and more \blue{customers} , selfishly , would complicate things for me . \blue{chicken} panang is quite good with a superb brown \blue{rice} . \blue{decent} wine \blue{list} . after three visits the wait staff remembered what i like ( complicated )  and always get the order right . \\
\cmidrule[\heavyrulewidth]{1-2}
CGMH &  they \blue{hesitate} to \blue{give} \blue{customers} their \blue{chicken} fried \blue{rice} and a \blue{decent} wine \blue{list} .\\
\cmidrule[\heavyrulewidth]{1-2}
NMSTG &  they \blue{hesitate} to an wonderful time to \blue{give} it about a table , love the \blue{customers} \blue{chicken} \blue{rice} and dishes seafood and \blue{decent} at the \blue{list} .\\
\cmidrule[\heavyrulewidth]{1-2}
\textsc{Pointer} (Greedy, base) & i just did not even \blue{hesitate} to admit , i should \blue{give} credit cards to my \blue{customers} here . the beijing \blue{chicken} and fried \blue{rice} were spot on , a \blue{decent} side on my favorite \blue{list} . \\
\cmidrule[\heavyrulewidth]{1-2}
\textsc{Pointer} (ILBS, base) & i don't have to \blue{hesitate} that they should \blue{give} five stars . i will be one of their repeat \blue{customers} . like the basil \blue{chicken} and basil fried \blue{rice} , it was \blue{decent} on my \blue{list} . \\
\cmidrule[\heavyrulewidth]{1-2}
\textsc{Pointer} (Greedy, Large) & service is very slow , don ' t \blue{hesitate} to tell manager to \blue{give} some feedbacks as their job is to take care of their \blue{customers} . had the vegetable medley soup and \blue{chicken} . both were cooked well . the garlic \blue{rice} did not have the vegetable and was fairly \blue{decent} . they are changing the flavor and \blue{list} of menu items . \\
\cmidrule[\heavyrulewidth]{1-2} 
Wiki zero-shot& he did not \blue{hesitate} himself to \blue{give} it to his \blue{customers} , such as \blue{chicken} , and steamed \blue{rice} , a very \blue{decent} item on the \blue{list} . \\
\cmidrule[\heavyrulewidth]{1-2} 
\end{tabular}
\end{table}

\begin{table}[H]
\begin{tabular}{p{0.5in}|p{2.2in} }
\cmidrule[\heavyrulewidth]{1-2}
Keywords & good potential bad maintained replaced dirty disgusting \\
\cmidrule[\heavyrulewidth]{1-2} 
\textsc{Oracle} & has \blue{good} \blue{potential} but very \blue{bad} \blue{maintained} . the padding is done , needs to be \blue{replaced} , holes everywhere . so are those huge flowers or what ever those are . ripped . very \blue{dirty} too . there was a a very dirty towel laying on the floor \blue{disgusting} . please the city of vegas come and clean it ! \\
\cmidrule[\heavyrulewidth]{1-2}
CGMH &  \blue{good} \blue{potential} but \blue{bad} service. not \blue{maintained} . it \blue{replaced} a \blue{dirty} box . \blue{disgusting} .\\
\cmidrule[\heavyrulewidth]{1-2}
NMSTG &  do a \blue{good} price . not like the and \blue{potential} \blue{bad} \blue{maintained} has disgusting . \blue{replaced} been , \blue{dirty} and \blue{disgusting} .\\
\cmidrule[\heavyrulewidth]{1-2}
\textsc{Pointer} (Greedy, base) & the food was very \blue{good} . it really has more \blue{potential} maybe , but it smells really \blue{bad} . its not very well \blue{maintained} either . trash cans were \blue{replaced} only when they were \blue{dirty} . the floors were utterly \blue{disgusting} . \\
\cmidrule[\heavyrulewidth]{1-2}
\textsc{Pointer} (ILBS, base) & the food is really \blue{good} . this location has \blue{potential} to be pretty \blue{bad} and not very well \blue{maintained} when it was \blue{replaced} , its super \blue{dirty} , just plain \blue{disgusting} . \\
\cmidrule[\heavyrulewidth]{1-2}
\textsc{Pointer} (Greedy, Large) & this gym is not so \blue{good} . overall it has a lot of \blue{potential} for being better but it is too \blue{bad} that it is not clean and un \blue{maintained} and towels are in desperate need to be \blue{replaced} regularly . the floors are very \blue{dirty} and the higher floors have become filthy \blue{disgusting} when i visited here . \\
\cmidrule[\heavyrulewidth]{1-2} 
Wiki zero-shot& it is \blue{good} it has no \blue{potential} , and the \blue{bad} taste can be \blue{maintained} until they are \blue{replaced} by a \blue{dirty} , and \blue{disgusting} one . \\
\cmidrule[\heavyrulewidth]{1-2} 
\end{tabular}
\end{table}

\begin{table}[H]
\begin{tabular}{p{0.5in}|p{2.2in} }
\cmidrule[\heavyrulewidth]{1-2}
Keywords & love animal style long line expected quick \\
\cmidrule[\heavyrulewidth]{1-2} 
\textsc{Oracle} & who doesn t \blue{love} in and out . \blue{animal} style is a must . \blue{long} \blue{line} but \blue{expected} , it goes \blue{quick} anyways so don t let that discourage you . \\
\cmidrule[\heavyrulewidth]{1-2}
CGMH & \blue{love} this place . \blue{animal} \blue{style} food . \blue{long} \blue{line} than \blue{expected} for \blue{quick} .\\
\cmidrule[\heavyrulewidth]{1-2}
NMSTG &  \blue{love} \blue{animal} chicken . it was \blue{style} \blue{long} a bit so good . the \blue{line} is it was even on on a time and we \blue{expected} to go but \blue{quick} .\\
\cmidrule[\heavyrulewidth]{1-2}
\textsc{Pointer} (Greedy, base) & great little breakfast spot . i \blue{love} having the double with \blue{animal} \blue{style} fries and protein style etc . have a super \blue{long} wait \blue{line} , but its just as \blue{expected} and it always moves pretty \blue{quick} too . \\
\cmidrule[\heavyrulewidth]{1-2}
\textsc{Pointer} (ILBS, base) & y all you just gotta \blue{love} about this place is the double \blue{animal} \blue{style} and protein style . it was a \blue{long} \blue{line} , but i \blue{expected} it to be \blue{quick} . \\
\cmidrule[\heavyrulewidth]{1-2}
\textsc{Pointer} (Greedy, Large) & great burger and good price . i \blue{love} that they have non chain locations . i like the \blue{animal} \blue{style} fries too . have to wait \blue{long} as there is always traffic but the \blue{line} can be much shorter than i had \blue{expected} and they are always send out pretty \blue{quick} . very impressed ! \\
\cmidrule[\heavyrulewidth]{1-2} 
Wiki zero-shot& he also has \blue{love} with the \blue{animal} and his \blue{style} , and was \blue{long} as the finish \blue{line} , and was \blue{expected} to be \blue{quick} . \\
\cmidrule[\heavyrulewidth]{1-2} 
\end{tabular}
\end{table}

\begin{table}[H]
\begin{tabular}{p{0.5in}|p{2.2in} }
\cmidrule[\heavyrulewidth]{1-2}
Keywords & great great service happy found close home \\
\cmidrule[\heavyrulewidth]{1-2} 
\textsc{Oracle} & \blue{great} sushi and \blue{great} \blue{service} . i m really \blue{happy} to have \blue{found} a good sushi place so \blue{close} to \blue{home} ! \\
\cmidrule[\heavyrulewidth]{1-2}
CGMH &  \blue{great} price and \blue{great} customer \blue{service} . very \blue{happy} that i \blue{found} this place \blue{close} to my \blue{home} .\\
\cmidrule[\heavyrulewidth]{1-2}
NMSTG &  \blue{great} food and \blue{great} \blue{service} . a \blue{happy} and \blue{found} a year in \blue{close} for them . keep them \blue{home} here .\\
\cmidrule[\heavyrulewidth]{1-2}
\textsc{Pointer} (Greedy, base) & amazing food . \blue{great} quality food . \blue{great} prices and friendly \blue{service} staff . so \blue{happy} and surprised to have finally \blue{found} such a wonderful nail salon so \blue{close} to my work and \blue{home} . \\
\cmidrule[\heavyrulewidth]{1-2}
\textsc{Pointer} (ILBS, base) & this is just \blue{great} food . \blue{great} food and wonderful \blue{service} . very \blue{happy} to have finally \blue{found} a chinese restaurant \blue{close} to my \blue{home} . \\
\cmidrule[\heavyrulewidth]{1-2}
\textsc{Pointer} (Greedy, Large) & wow . i have been here twice . \blue{great} times here . food always has been \blue{great}  and the customer \blue{service} was wonderful . i am very \blue{happy} that we finally \blue{found} our regular pad thai restaurant that is \blue{close} to where we work now and our \blue{home} . pleasantly surprised ! \\
\cmidrule[\heavyrulewidth]{1-2} 
Wiki zero-shot& he was a \blue{great} teacher and a \blue{great} love of the \blue{service} he was very \blue{happy} , and he \blue{found} himself in the \blue{close} to his \blue{home} . \\
\cmidrule[\heavyrulewidth]{1-2} 
\end{tabular}
\end{table}

\begin{table*}[ht!]
\small
\centering
\begin{tabular}{r H  r H  r | H r  H r | r| H H H r  |r  r | r |r}
	\cmidrule[\heavyrulewidth]{1-18}
	 & \multicolumn{4}{c|}{NIST} & \multicolumn{4}{c|}{BLEU} & METEOR & \multicolumn{4}{c|}{Entropy} & \multicolumn{2}{c|}{Dist} & \multicolumn{1}{c|} {PPL} & \multicolumn{1}{c} {Avg Len}  \\ 
	Method & N-1 & N-2 & N-3 & N-4 & B-1 & B-2 & B-3 & B-4 &  & E-1 & E-2 & E-3 & E-4 &  D-1 &D-2 & & \\
	\cmidrule[\heavyrulewidth]{1-18} 
	Greedy (+Wiki) & 2.84 & 3.04 & 3.06 & 3.06 & 33.73\% & 13.01\% & 5.56\% & 2.51\% & \textbf{16.38\%} & 6.14 & 9.19 & 10.09 & 10.22 & 11.10\% & 57.78\% & \textbf{56.7} & 31.32\\
	ILBS (+Wiki) & 2.97 & 3.20 & 3.22 & 3.22 & 32.69\% & 14.00\% & 6.24\% & 2.99\% & 15.71\% & 6.05 & 8.97 & 9.77 & 9.86 & 13.17\% & 61.22\% & 66.4 & 22.59\\
	Greedy (+Wiki,L) & 3.04 & \textbf{3.28} & \textbf{3.29} & \textbf{3.30} & 35.75\% & \textbf{14.04\%} & 6.10\% & \textbf{3.04\%} & 15.90\% & 6.15 & 9.17 & 9.97 & 10.09 & 12.23\% & 60.86\% & \textbf{54.7} & 27.99\\
	\cmidrule[\heavyrulewidth]{1-18} 
	Wiki zero-shot & 2.59 & 2.80 & 2.81 & 2.82 & 25.94\% & 11.38\% & 4.85\% & 1.84\% & 15.12\% & 5.69 & 8.47 & 9.55 & 9.73 & 14.33\% & 53.97\% & 62.9 & 20.68\\
	\cmidrule[\heavyrulewidth]{1-18} 
	Human & - & - & - & - & -& - & - & - & - & 6.17 & 9.20 & 9.93 & 10.05 & 11.80\% & 62.44\% & 47.4 & 27.85\\
	\cmidrule[\heavyrulewidth]{1-18}
	\end{tabular}
\caption{Additional evaluation results on the News dataset. ILBS denotes beam search. ``+Wiki'' denotes fine-tuning on the Wiki-pretrained model.  ``Human'' represents the held-out human reference. ``Wiki zero-shot'' represents zero-shot generation from the pre-trained model.}
\vspace{-3mm}
\label{tab:wiki_news}
\end{table*}

\begin{table*}[ht!]
\small
\centering
\begin{tabular}{r H  r H  r | H r  H r | r| H H H r  |r  r | r |r}
	\cmidrule[\heavyrulewidth]{1-18}
	 & \multicolumn{4}{c|}{NIST} & \multicolumn{4}{c|}{BLEU} & METEOR & \multicolumn{4}{c|}{Entropy} & \multicolumn{2}{c|}{Dist} & \multicolumn{1}{c|} {PPL} & \multicolumn{1}{c} {Avg Len}  \\ 
	Method & N-1 & N-2 & N-3 & N-4 & B-1 & B-2 & B-3 & B-4 &  & E-1 & E-2 & E-3 & E-4 &  D-1 &D-2 & & \\
	\cmidrule[\heavyrulewidth]{1-18} 
	Greedy (+Wiki) & 3.02 & 3.27 & 3.30 & 3.30 & 38.25\% & 15.63\% & 6.93\% & 3.32\% & 16.14\% & 5.96 & 9.18 & 10.37 & 10.64 & 7.51\% & 46.12\% & 71.9 & 48.22\\
	ILBS (+Wiki) & 3.08 & 3.34 & 3.39 & 3.38 & 39.63\% & 16.68\% & 7.62\% & 3.65\% & 15.57\% & 5.96 & 9.06 & 10.08 & 10.44 & 9.43\% & 50.66\% & 61.0 & 35.18\\
	Large (+Wiki) & 3.19 & \textbf{3.49} & \textbf{3.52} & \textbf{3.53} & 39.41\% & \textbf{16.78\%} & 7.71\% & \textbf{3.79\%} & 16.69\% & 5.79 & 8.89 & 10.19 & 10.56 & 6.94\% & 41.2\% & \textbf{55.5} & 48.05\\
	\cmidrule[\heavyrulewidth]{1-18}
	Wiki zero-shot & 0.81 & 0.86 & 0.87 & 0.87 & 21.75\% & 8.56\% & 3.20\% & 1.30\% & 12.85\% & 5.34 & 8.08 & 9.48 & 9.90 & 10.09\% & 41.97\% & 62.9 & 26.80\\
	\cmidrule[\heavyrulewidth]{1-18}
	Human & - & - & - & - & -& - & - & - & - & 6.20 & 9.45 & 10.49 & 10.70 & 10.67\% & 52.57\% & 55.4 & 50.36\\
	\cmidrule[\heavyrulewidth]{1-18}
	\end{tabular}
\caption{Additional evaluation results on the Yelp dataset. ILBS denotes beam search. ``+Wiki'' denotes fine-tuning on the Wiki-pretrained model.  ``Human'' represents the held-out human reference. ``Wiki zero-shot'' represents zero-shot generation from the pre-trained model.}
\vspace{-3mm}
\label{tab:wiki_yelp}
\end{table*}

\begin{table*}[ht!]
\scriptsize
\centering

\begin{tabular}{r r | r| r l|r r | r| r l}
\cmidrule[\heavyrulewidth]{1-10}
 \multicolumn{10}{c}{\textbf{Informativeness}: \textit{A and B, which is more semantically meaningful and consistent?}}\\
 \cmidrule[\heavyrulewidth]{1-10} 
  \multicolumn{5}{c|}{News dataset} & \multicolumn{5}{c}{Yelp dataset} \\
\cmidrule[\heavyrulewidth]{1-10} 
\multicolumn{2}{c|}{System A} & Neutral & \multicolumn{2}{c|}{System B} & \multicolumn{2}{c|}{System A} & Neutral & \multicolumn{2}{c}{System B} \\ 
\cmidrule[\heavyrulewidth]{1-10}
\cmidrule{1-10}
\textsc{Pointer}(large)& 35.4\% & 27.7\% & \bf{36.9} \% & \textsc{Pointer}(base) &
\textsc{Pointer}(large)& \bf{41.4}\% & 26.6\% & 32.1 \% & \textsc{Pointer}(base) ***\\
\cmidrule{1-10} 
\textsc{Pointer}(large)& 20.3\% & 22.7\% & \textbf{57.1}\% & Human *** &
\textsc{Pointer}(large)& 27.2\% & 24.4\% & \textbf{48.5}\% & Human *** \\
\cmidrule[\heavyrulewidth]{1-10}
%
%
\cmidrule[\heavyrulewidth]{1-10}
 \multicolumn{10}{c}{\textbf{Fluency}: \textit{A and B, which is more grammatical and fluent?}}\\
 \cmidrule[\heavyrulewidth]{1-10} 
  \multicolumn{5}{c|}{News dataset} & \multicolumn{5}{c}{Yelp dataset} \\
\cmidrule[\heavyrulewidth]{1-10} 
\multicolumn{2}{c|}{System A} & Neutral & \multicolumn{2}{c|}{System B} & \multicolumn{2}{c|}{System A} & Neutral & \multicolumn{2}{c}{System B} \\ 
\cmidrule[\heavyrulewidth]{1-10}
\cmidrule{1-10}
\textsc{Pointer}(large)& \bf{38.4}\% & 28.5\% & 33.2 \% & \textsc{Pointer}(base) &
\textsc{Pointer}(large)& \bf{41.1}\% & 28.1\% & 30.8 \% & \textsc{Pointer}(base) ***\\
\cmidrule{1-10} 
\textsc{Pointer}(large)& 16.7\% & 15.8\% & \textbf{67.5}\% & Human *** &
\textsc{Pointer}(large)& 27.1\% & 21.9\% & \textbf{51.1}\% & Human *** \\
\cmidrule[\heavyrulewidth]{1-10}
%
%
\cmidrule[\heavyrulewidth]{1-10}
 \multicolumn{10}{c}{\textbf{Informativeness}: \textit{A and B, which is more informative?}}\\
 \cmidrule[\heavyrulewidth]{1-10} 
  \multicolumn{5}{c|}{News dataset} & \multicolumn{5}{c}{Yelp dataset} \\
\cmidrule[\heavyrulewidth]{1-10} 
\multicolumn{2}{c|}{System A} & Neutral & \multicolumn{2}{c|}{System B} & \multicolumn{2}{c|}{System A} & Neutral & \multicolumn{2}{c}{System B} \\ 
\cmidrule[\heavyrulewidth]{1-10}
\cmidrule{1-10}
\textsc{Pointer}(large)& 32.1\% & 27.6\% & \bf{40.4} \% & \textsc{Pointer}(base) &
\textsc{Pointer}(large)& \bf{41.6}\% & 25.0 \% & 33.4 \% & \textsc{Pointer}(base) ***\\
\cmidrule{1-10} 
\textsc{Pointer}(large)& 31.9\% & 17.1\% & \textbf{51.0}\% & Human *** &
\textsc{Pointer}(large)& 35.9\% & 14.7\% & \textbf{49.4}\% & Human ***\\
\cmidrule[\heavyrulewidth]{1-10}
\end{tabular}

	\vspace{-2mm}
\caption{{\bf Human Evaluation} on two datasets for semantic consistency, fluency and informativeness, showing preferences (\%) for our \textsc{Pointer}(large) model vis-a-vis \textsc{Pointer}(base) model and real human responses. Numbers in bold indicate the most preferred systems. Significant differences (p $\leq$ 0.001) are indicated as ***. 
\vspace{-3mm}
}\label{tab:human_eval_full}

\end{table*}

\section{Additional Human Evaluation information and Results}
\label{app:human_eval_full}
There were 145 judges in all: 5 judges evaluated each pair of outputs to be reasonably robust against spamming. P-values are all p<0.00001 (line 721), computed using 10000 bootstrap replications. Judges were lightly screened by our organization for multiple screening tasks. 

We present the additional human evaluation results on \textsc{Pointer} large model vs base model in table~\ref{tab:human_eval_full}. In general, for the news dataset the results are mixed. For the yelp dataset, the large model wins with a large margin. All results are still far away from the human oracle in all three aspects.

\section{Additional Automatic Evaluation Results}
\label{app:wiki}
We provide the full evaluation result data including Wikipedia zero-shot learning results in Table~\ref{tab:wiki_news} and Table~\ref{tab:wiki_yelp}. Note that zero-shot generations from Wikipedia pre-trained model yield the lowest perplexity, presumably because the Wikipedia dataset is large enough so that the model trained on it can learn language variability, thus delivering fluent generated results.

\section{Inference Details}
During inference time, we use a decaying schedule to discourage the model from generating non-interesting tokens, including $\mathtt{[NOI]}$ and some other special tokens, punctuation and stop words. To do this, we use a decay multiplier $\eta$ on the logits of these tokens before computing the softmax. The $\eta$ is set to be $\eta=\min(0.5 + \lambda * s)$, where $s$ is the current stage and $\lambda$ is an annealing hyper-parameter. In most of the experiments, $\lambda$ is set at $0.5$

\section{Human Evaluation Template}
\label{app:human}
\begin{figure*}
	\includegraphics[width=0.99\linewidth]{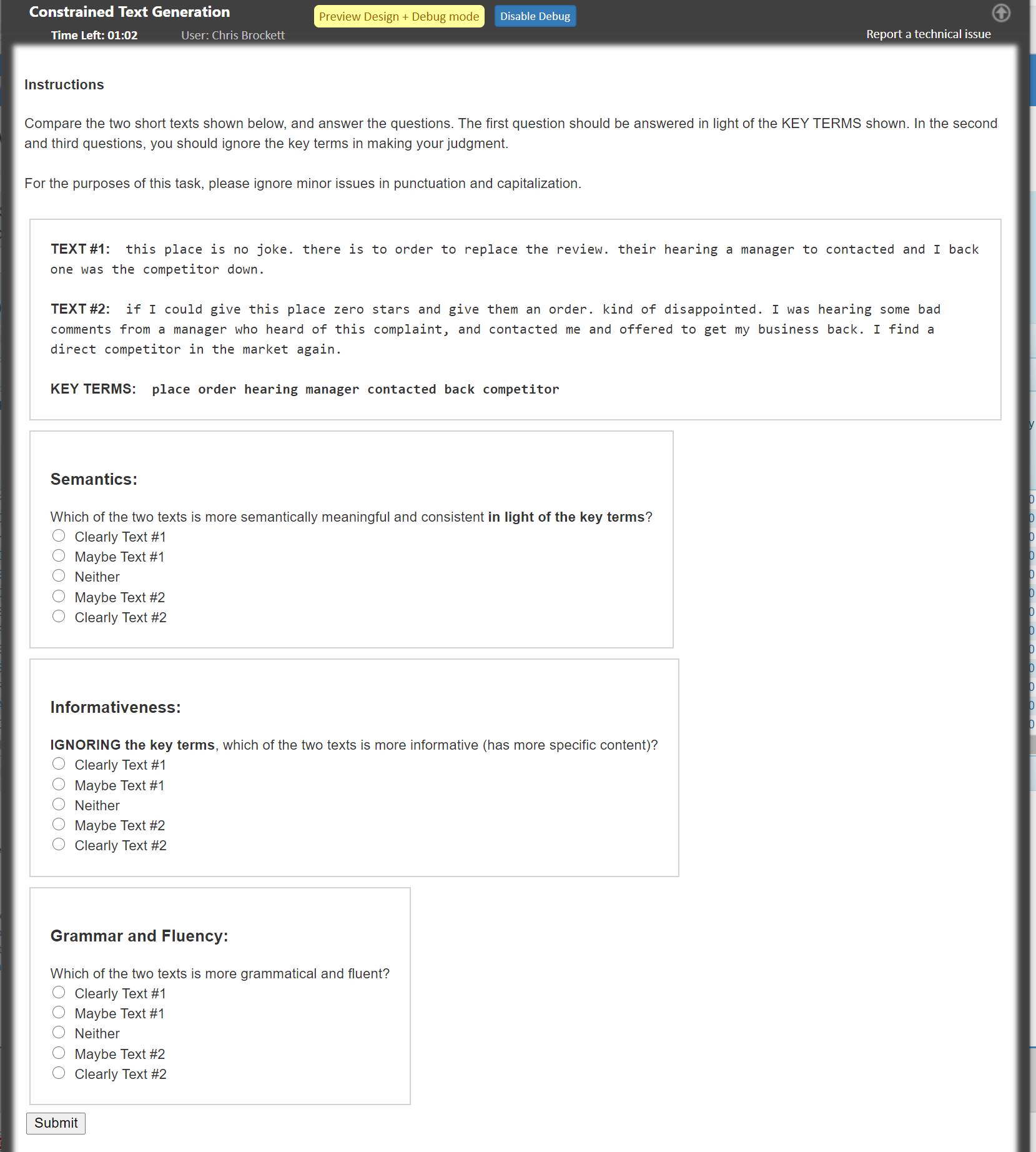}
	\caption{Human evaluation template.} 
	\label{fig:human}
	\vspace{-3mm}
\end{figure*}

See Figure~\ref{fig:human} for human evaluation template 

\end{document}